\documentclass{article}

\usepackage[preprint]{neurips_2026}

\usepackage[utf8]{inputenc} 
\usepackage[T1]{fontenc}    
\usepackage{url}    
\usepackage{booktabs}   
\usepackage{amsmath}
\usepackage{amsfonts}       
\usepackage{nicefrac}       
\usepackage{microtype}      
\usepackage{xcolor}         
\usepackage{graphicx}
\usepackage{multirow}
\usepackage{pdflscape}
\usepackage{caption}
\usepackage{tcolorbox}
\usepackage{pifont}
\usepackage{xcolor}
\usepackage{tcolorbox}
\tcbuselibrary{skins}
\usepackage{tabularx}
\usepackage{fontawesome5}
\usepackage{enumitem}
\usepackage{array}
\usepackage[table]{xcolor}
\usepackage{array}
\usepackage{graphicx}
\usepackage{booktabs}
\usepackage{subcaption}
\usepackage{wasysym}
\usepackage{pifont}
\usepackage{float}

\definecolor{uclablue}{rgb}{0.15, 0.45, 0.68}

\usepackage[
    pagebackref,
    breaklinks,
    colorlinks=true,
    citecolor=uclablue,
    linkcolor=uclablue,
    urlcolor=uclablue
]{hyperref}
\title{\textsc{ESI-Bench}: Towards Embodied Spatial Intelligence that Closes the Perception-Action Loop}

\author{%
\vspace{-1em}
\footnotesize
Yining Hong*$^{1}$ \;
Jiageng Liu*$^{2}$ \;
Han Yin$^{1}$ \;
Manling Li$^{3}$ \;
Leonidas Guibas$^{1}$ \;
Li Fei-Fei$^{1}$ \;
Jiajun Wu$^{1}$ \;
Yejin Choi$^{1}$ \\
\\
\footnotesize
$^{1}$Stanford University \;
$^{2}$UCLA \;
$^{3}$Northwestern University\\
\url{https://esi-bench.github.io/}
}

\begin{document}

\definecolor{esiOrange}{RGB}{200,120,40}
\definecolor{esiBlue}{RGB}{70,120,200}
\definecolor{esiPurple}{RGB}{140,90,180}
\definecolor{esiGreen}{RGB}{60,140,120}
\definecolor{orangeBg}{RGB}{255,245,235}
\definecolor{blueBg}{RGB}{235,242,255}
\definecolor{purpleBg}{RGB}{245,238,255}
\definecolor{greenBg}{RGB}{235,250,245}

\maketitle
\vspace{-2em}
\begin{figure}[h]
\centering
\includegraphics[width=\linewidth]{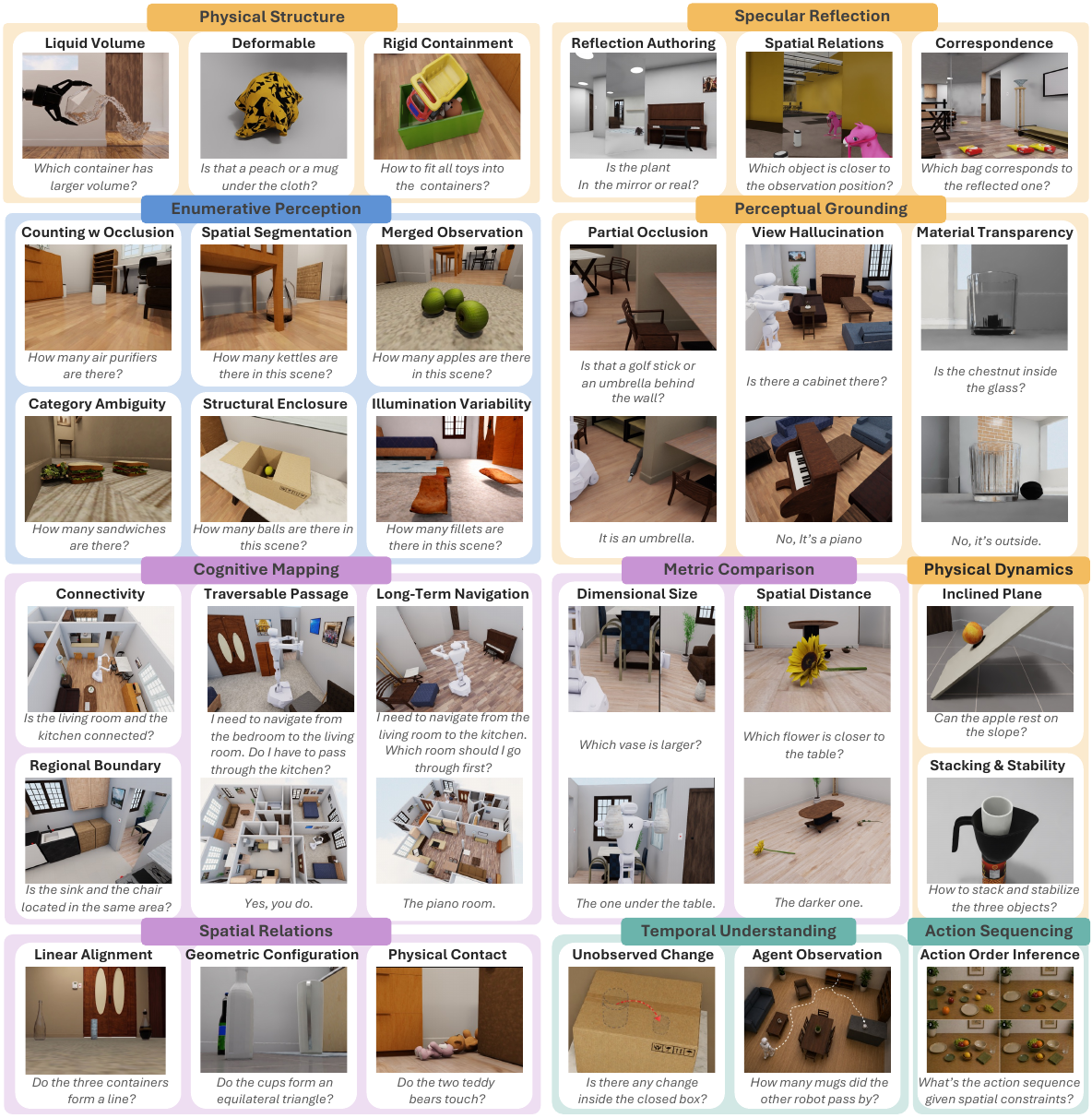}
\caption{\small \textbf{\textsc{ESI-Bench}} is a comprehensive benchmark for embodied spatial intelligence, spanning 10 task categories and 29 subcategories organized around Spelke's four core knowledge systems \citep{spelke2007core}: \textcolor{esiOrange}{object representation}, \textcolor{esiPurple}{layout and geometry}, \textcolor{esiBlue}{number representation}, and \textcolor{esiGreen}{agents and goal-directed actions}. }
\label{fig:teaser}
\end{figure}

\begin{abstract}
\vspace{-0.5em}
Spatial intelligence unfolds through a perception–action loop: agents act to acquire observations, and reason about how observations vary as a function of action. 
Rather than passively processing what is \textit{seen}, they actively uncover what is \textit{unseen} - occluded structure, dynamics, containment, and functionality that cannot be resolved from passive sensing alone.
We move beyond prior formulations of spatial intelligence that assume oracle observations by recasting the observer as an actor.
 We introduce \textsc{ESI-Bench}, a comprehensive benchmark for embodied spatial intelligence spanning 10 task categories and 29 subcategories built on OmniGibson, grounded in Spelke’s core knowledge systems. Agents must decide what abilities to deploy — perception, locomotion, and manipulation — and how to sequence them to actively accumulate task-relevant evidence.
We conduct extensive experiments on state-of-the-art MLLMs and find that active exploration substantially outperforms passive counterparts, with agents spontaneously discovering emergent spatial strategies without explicit instructions, while random multi-view often adds noise rather than signal despite consuming far more images. Most failures stem not from weak perception but from action blindness: poor action choices lead to poor observations, which in turn drive cascading errors. While explicit 3D grounding stabilizes reasoning on depth-sensitive tasks, imperfect 3D representation proves more harmful than 2D baselines by distorting spatial relations. Human studies further reveal that unlike humans who seek falsifying viewpoints and revise beliefs under contradiction, models commit prematurely with high confidence regardless of evidence quality, exposing a metacognitive gap that neither better perception nor more embodied interaction alone can close. All data, dataset construction codes, and evaluation scripts are publicly available at \url{https://esi-bench.github.io/}.
\end{abstract}

\vspace{-1em}
\section{Introduction}
\vspace{-0.5em}
Perception is often characterized in cognitive science as \textit{perceptually guided action} \citep{varela1991embodied, gibson1979ecological}: a perception–action loop where knowing how observations change as a function of action \citep{oregan2001sensorimotor}, and knowing which actions elicit informative sensing, are often more challenging than sensing itself. This is especially critical in spatial intelligence, which concerns not only what is \textit{seen}, but also what is \textit{unseen}. Latent physical properties such as occluded structure, dynamics, containment, and functionality that are inaccessible to passive sensing must be actively revealed through interaction, making spatial intelligence inherently embodied.


We move beyond prior formulations of spatial intelligence that assume passive oracle observations \citep{liu2023visualspatialreasoning, yang2025thinkingspacemultimodallarge, yang2025mmsibenchbenchmarkmultiimagespatial} by recasting the observer as an actor.
Our work contrasts with prior works in three key ways: (1) \textit{from spatial sensing to spatial competence}, where agents are evaluated not only on what they can perceive, but on whether they know which embodied abilities to deploy to solve spatial tasks; (2) \textit{selective sensing}, where agents must determine which observations are worth acquiring, prioritizing task-relevant information over redundant or uninformative inputs; and (3) \textit{resolving perceptual ambiguities}, where agents must reason through misleading observations to infer hidden spatial structures and underlying physical constraints beyond what is directly observed.



We introduce \textsc{ESI-Bench}, a comprehensive benchmark for embodied spatial intelligence spanning 10 task categories, 29 subcategories and 3,081 task instances, addressing the critical perception–action gap by focusing on questions that cannot be answered from passive observation alone. Our category design follows Spelke's core knowledge systems \citep{spelke2007core}, which identify four faculties of spatial intelligence: object representation, layout and geometry, number representation, and agents \& goal-driven actions. Building on these theoretical foundations, we conduct human surveys to identify the most challenging spatial tasks that require embodied interaction and manipulation within each faculty, which are distilled into a structured taxonomy spanning diverse forms of spatial reasoning, as illustrated in Figure \ref{fig:teaser}. These tasks only become meaningful when an agent has a body, a belief state, and physical stakes in the outcome: the agent must determine what abilities to deploy (perception, locomotion, manipulation), which actions to take (where to move, what to probe, how to manipulate), and how to execute them in the right order to successfully answer the questions.



We conduct extensive experiments on state-of-the-art MLLMs across three paradigms: passive single-view, passive random multi-view, and active exploration, alongside a ground-truth oracle that separates perception errors from action errors. Our experiments reveal multiple key insights: (1) \textit{active exploration unlocks emergent spatial strategies}: without explicit instruction, active agents spontaneously discover diverse action compositions, driving substantial gains over passive counterparts while passive random multi-view, despite consuming far more images, often adds noise rather than signal; (2) \textit{action blindness dominates perceptual blindness}: for most tasks the bottleneck is not perception but action selection, as models given oracle viewpoints succeed largely, yet certain tasks expose a hard perceptual ceiling that no action can overcome; and (3) \textit{failures cascade and compound}: suboptimal actions produce uninformative views, which trigger worse subsequent actions, creating a compounding chain of errors that cannot be recovered within the step budget.

We further investigate whether explicit 3D representations can help. We find that while explicit 3D grounding stabilizes reasoning on depth-sensitive tasks by recovering information that 2D projections fundamentally lose, imperfect reconstructions prove more harmful than 2D baselines, as geometric artifacts actively distort fine-grained spatial relations and mislead downstream reasoning.
Finally, human studies expose a critical gap in epistemic calibration (\textit{i.e.}, a model’s confidence matches the quality and uncertainty of its evidence). We observe that unlike humans who actively seek falsifying viewpoints, explore orthogonal angles, and revise their beliefs when contradicted, models commit prematurely with uniformly high confidence regardless of evidence quality, anchoring to first impressions and ignoring contradictory observations, a metacognitive failure that neither better perception nor more embodied interaction alone can close.
\section{Related Works}
\vspace{-1.5em}
\begin{table}[h]
\centering
\footnotesize
\caption{Comparison with spatial reasoning and  embodied benchmarks.
Here, \textit{embodied} means that an agent is situated in an embodied environment and
acts through an explicit action space in the environment. \textbf{Active?}: whether the benchmark supports active
perception. \textbf{Action Space}: L = locomotion,
P = perception, M = manipulation. \textbf{Core Knowledge}: Obj = object, Geom = geometry,
Num = number, Phys = physics, Agt = agent. \textbf{Hidden}: whether the
benchmark targets latent properties such as occlusion, containment, dynamics that are inaccessible from passive observation.}
\label{tab:benchmark_comparison}
\setlength{\tabcolsep}{2.1pt}
\renewcommand{\arraystretch}{0.9}

\begin{tabular}{@{}lccccc@{}}
\toprule
\textbf{Benchmark} & \textbf{Type} & \textbf{Active?} & \textbf{Action} & \textbf{Hidden} & \textbf{Core} \\
\midrule
VSR~\citep{liu2023visualspatialreasoning}
& Spatial & \ding{55} & P & \ding{55} & Geom \\

BLINK~\citep{fu2024blinkmultimodallargelanguage}
& Spatial & \ding{55} & P & \ding{55} & Obj, Geom, Num \\

3DSRBench~\citep{ma20253dsrbenchcomprehensive3dspatial}
& Spatial & \ding{55} & P & \ding{55} & Obj, Geom \\

VSI-Bench~\citep{yang2025thinkingspacemultimodallarge}
& Spatial & \ding{55} & P & \ding{55} & Obj, Geom, Num \\

MMSI-Bench~\citep{yang2025mmsibenchbenchmarkmultiimagespatial}
& Spatial & \ding{55} & P & \ding{55} & Obj, Geom \\

MindCube~\citep{wang2026mindcubespatialmentalmodeling}
& Spatial & \ding{55} & P & \ding{51} & Geom, Phys \\

PhysBench~\citep{chow2025physbenchbenchmarkingenhancingvisionlanguage}
& Spatial & \ding{55} & P & \ding{51} & Obj, Geom, Phys \\

CausalSpatial~\citep{ma2026causalspatialbenchmarkobjectcentriccausal}
& Spatial & \ding{55} & P & \ding{51} & Obj, Phys \\

EmbSpatial-Bench~\citep{du2024embspatialbenchbenchmarkingspatialunderstanding}
& Spatial & \ding{55} & P & \ding{55} & Obj, Geom \\

OpenEQA~\citep{OpenEQA2023}
& Embodied & \ding{51} & L, P & \ding{55} & Obj, Geom \\

EmbodiedBench~\citep{yang2025embodiedbenchcomprehensivebenchmarkingmultimodal}
& Embodied & \ding{51} & L, P, M & \ding{55} & Obj, Geom, Agt \\

EmbodiedEval~\citep{cheng2025embodiedevalevaluatemultimodalllms}
& Embodied & \ding{51} & L, P, M & \ding{55} & Obj, Geom, Agt \\

EXPRESS-Bench~\citep{jiang2025destinationnovelbenchmarkexplorationaware}
& Embodied & \ding{51} & L, P & \ding{55} & Obj, Geom \\

ESPIRE~\citep{zhao2026espirediagnosticbenchmarkembodied}
& Spatial+Embodied & \ding{55} & P, M & \ding{55} & Obj, Geom \\

CHAIN~\citep{wu2026perceptionactioninteractivebenchmark}
& Embodied & \ding{51} & P, M & \ding{51} & Obj, Geom, Phys \\
 
\midrule
\textbf{\textsc{ESI-Bench} (ours)}
& \textbf{Spatial+Embodied} & \ding{51} & \textbf{L, P, M} & \ding{51}
& \textbf{Obj, Geom, Num, Phys, Agt} \\
\bottomrule
\end{tabular}%
\end{table}

\vspace{-0.5em}
\textbf{Benchmarks for Spatial Reasoning.}
Evaluation of spatial intelligence has scaled rapidly, yet most benchmarks still assume fixed observations. VSR~\citep{liu2023visualspatialreasoning} evaluates spatial relations in single images; BLINK~\citep{fu2024blinkmultimodallargelanguage} and 3DSRBench~\citep{ma20253dsrbenchcomprehensive3dspatial} extend to visual perception and fine-grained 3D reasoning; and SpatialScore~\citep{wu2026spatialscorecomprehensiveevaluationspatial} unifies VGBench and other datasets with a tool-augmented agent. Recent benchmarks broaden inputs to egocentric video in VSI-Bench~\citep{yang2025thinkingspacemultimodallarge}, multi-image consistency in MMSI-Bench~\citep{yang2025mmsibenchbenchmarkmultiimagespatial}, partial-observation mental modeling in MindCube~\citep{wang2026mindcubespatialmentalmodeling}, long-horizon recall and continual counting in Cambrian-S/VSI-SUPER~\citep{yang2025cambriansspatialsupersensingvideo}, and latent physical structure and object-centric dynamics in PhysBench~\citep{chow2025physbenchbenchmarkingenhancingvisionlanguage} and CausalSpatial~\citep{ma2026causalspatialbenchmarkobjectcentriccausal}. However, even with richer inputs---images, views, videos, partial coverage, and dynamics---the observation process remains fixed, limiting diagnosis of active information acquisition. \textsc{ESI-Bench} keeps this diagnostic focus while making observation utility depend on the model's own decisions. Table~\ref{tab:benchmark_comparison} compares \textsc{ESI-Bench} with prior spatial intelligence benchmarks.

\vspace{-0.3em}
\textbf{Spatial Reasoning Methods in MLLMs.}
Recent MLLMs improve spatial understanding, but largely assume fixed observations. Geometry-based methods inject 3D priors into 2D backbones: SpatialVLM~\citep{Chen_2024_CVPR} uses synthesized 3D annotations and metric-depth supervision; SpatialBot~\citep{cai2024spatialbot} uses RGB-D and depth-oriented QA; SpatialRGPT~\citep{cheng2024spatialrgptgroundedspatialreasoning} uses 3D scene graphs and a depth plugin; and Spatial-MLLM~\citep{wu2025spatialmllmboostingmllmcapabilities} and VLM-3R~\citep{fan2026vlm3rvisionlanguagemodelsaugmented} add priors from monocular video or reconstructive tuning. Reasoning-based methods make inference explicit: SpatialCoT~\citep{liu2025spatialcotadvancingspatialreasoning} grounds CoT in spatial coordinates, while VILASR~\citep{wu2025reinforcingspatialreasoningvisionlanguage} combines textual reasoning with visual drawing. These methods improve reasoning over given inputs; \texttt{ESI-Bench} tests whether models can choose the observations they need.

\vspace{-0.3em}
\textbf{Embodied Evaluation and Active Perception.}
Another line evaluates models as embodied agents. EmbodiedBench~\citep{yang2025embodiedbenchcomprehensivebenchmarkingmultimodal} and EmbodiedEval~\citep{cheng2025embodiedevalevaluatemultimodalllms} measure navigation, interaction, and QA, while OpenEQA~\citep{OpenEQA2023} and EXPRESS-Bench~\citep{jiang2025destinationnovelbenchmarkexplorationaware} study embodied QA and exploration quality, with EAC discouraging disembodied reasoning. EmbSpatial-Bench~\citep{du2024embspatialbenchbenchmarkingspatialunderstanding} and ESPIRE~\citep{zhao2026espirediagnosticbenchmarkembodied} diagnose egocentric spatial reasoning, with ESPIRE separating localization from execution. Active perception methods learn observation selection, from human-demonstration learning in Vision in Action~\citep{xiong2025visionactionlearningactive} and humanoid head-rotation search in Thinking in 360$^{\circ}$~\citep{yu2025thinking360deghumanoidvisual} to viewpoint selection for VLA manipulation in SaPaVe~\citep{liu2026sapaveactiveperceptionmanipulation} and ActiveVLA~\citep{liu2026activevlainjectingactiveperception}. Closest to ours, CHAIN~\citep{wu2026perceptionactioninteractivebenchmark} evaluates closed-loop physical reasoning in mechanical puzzles, stacking, and packing. \texttt{ESI-Bench} complements these works with broader embodied spatial faculties across object, geometry, number, physics, and agent reasoning, including hidden states such as containment, occlusion, transparency, reflection, and unobserved scene change. Table~\ref{tab:benchmark_comparison} situates these differences.

\section{\textsc{ESI-Bench}}
\vspace{-0.5em}
In this section, we introduce ESI-Bench, a comprehensive benchmark comprising 10 task categories, 29 subcategories, and 3,081 task instances. We describe the benchmark setup and task construction 
pipeline; the full task taxonomy, per-category construction 
details, and human verification and generator-bias analysis 
are provided in Appendix~\ref{sec:task_description}, 
Appendix~\ref{sec:task_construction_details}, and 
Appendix~\ref{app:human_verification_generator_bias}, 
respectively.

\vspace{-0.5em}
\subsection{Benchmark Setup}
\begin{figure}[t]
    \centering

    \begin{subfigure}{\linewidth}
        \centering
        \includegraphics[width=\linewidth]{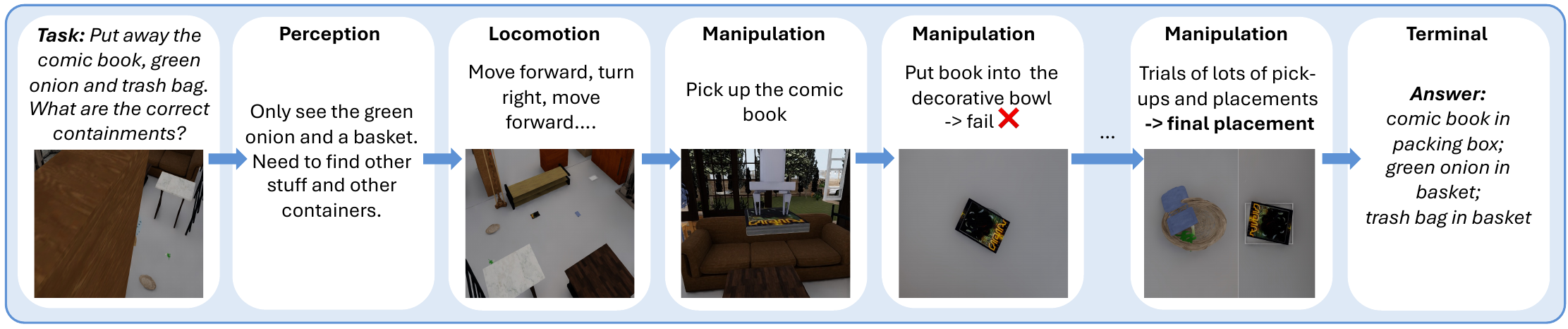}
        \vspace{-1.2em}
        \caption{A dataset example including question, answer and trajectory, from the Rigid Containment sub-category.}
        \label{fig:dataset_example}
    \end{subfigure}

    \vspace{3pt}

    \begin{subfigure}[c]{0.55\textwidth}
        \centering
        \scriptsize
        \setlength{\tabcolsep}{4pt}
        \renewcommand{\arraystretch}{0.9}
        \begin{tabular}{ll}
        \toprule
        \textbf{Action} & \textbf{Description} \\
        \midrule
        \multicolumn{2}{l}{\textit{Locomotion}} \\
        \texttt{move\_forward} / \texttt{move\_backward} & Translate agent along viewing axis \\
        \texttt{move\_left} / \texttt{move\_right}       & Translate agent laterally \\
        \texttt{move\_up} / \texttt{move\_down}           & Translate agent vertically \\
        \midrule
        \multicolumn{2}{l}{\textit{Perception}} \\
        \texttt{turn\_left} / \texttt{turn\_right}       & Rotate agent horizontally \\
        \texttt{turn\_up} / \texttt{turn\_down}           & Rotate agent vertically \\
        \midrule
        \multicolumn{2}{l}{\textit{Manipulation}} \\
        \texttt{pick\_up obj} & Pick up object \\
        \texttt{put obj inside obj} & Put object inside object \\
        \texttt{put obj on obj} & Put object on object \\
        \texttt{fill obj with water} & Fill object with water \\
        \texttt{pour obj from obj to obj} & Pour object from object to object \\
        \midrule
        \multicolumn{2}{l}{\textit{Terminal}} \\
        \texttt{answer($\hat{y}$, $c$)} & Commit to final answer with confidence $c$ \\
        \bottomrule
        \end{tabular}
        \caption{\textsc{ESI-Bench} agent action space.}
        \label{fig:action_space}
    \end{subfigure}
    \hfill
    \begin{subfigure}[c]{0.4\textwidth}
        \centering
        \includegraphics[width=\textwidth]{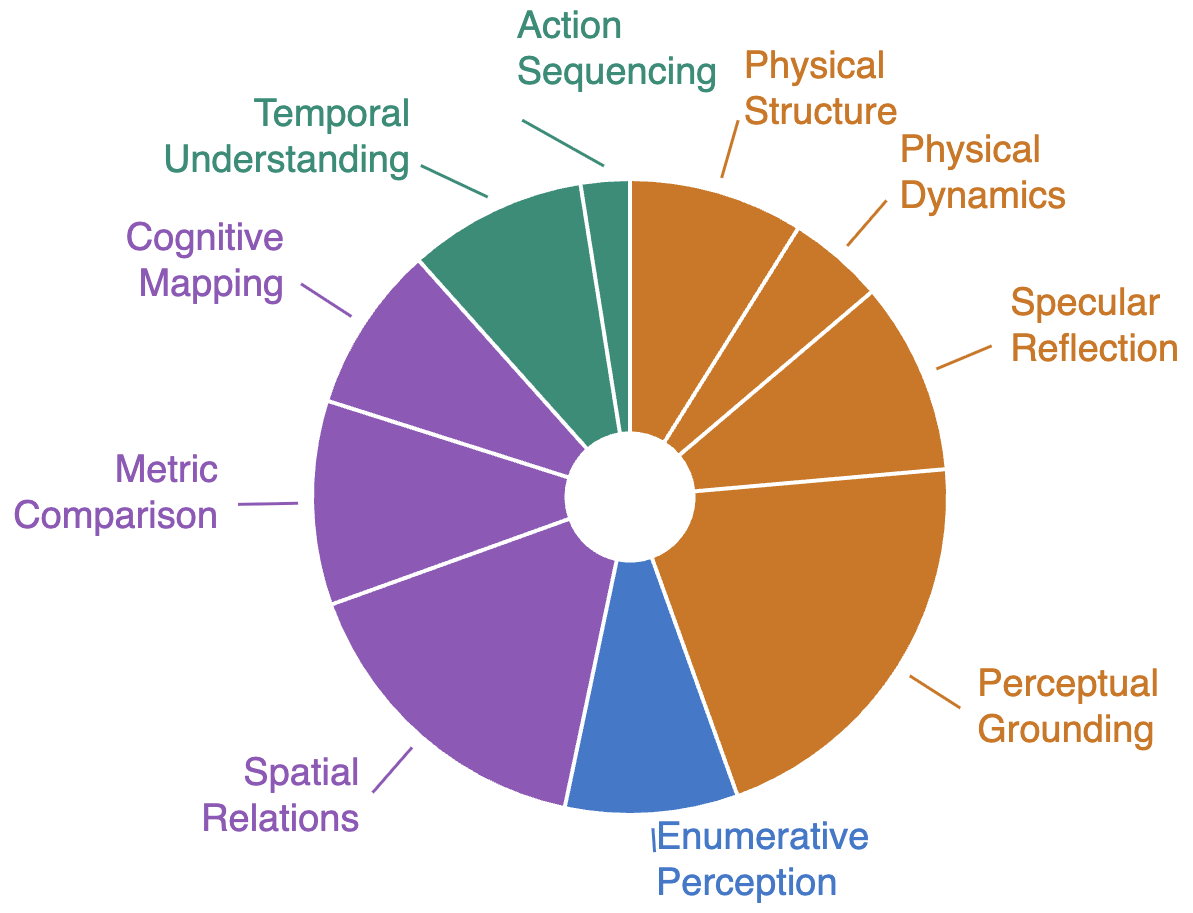}
        \caption{Distribution of task categories.}
        \label{fig:task_distribution}
    \end{subfigure}
    \vspace{-0.5em}
    \caption{
    Overview of \textsc{ESI-Bench}: dataset example, agent action space, and task distribution.
    }
    \label{fig:dataset_block}
    \vspace{-1.5em}
\end{figure}

\vspace{-0.5em}
\paragraph{Task Definition} Each task in ESI-Bench is defined as a tuple $(\mathcal{S}, p_0, q, y^*)$, where $\mathcal{S}$ is a 3D scene instantiated from the BEHAVIOR-1K scene pool with pre-loaded objects, $p_0$ is the agent's initial pose, $q$ is a natural-language question about a spatial property of the scene, and $y^*$ is the ground-truth answer. We formalize the environment as $\mathcal{E}=\langle \mathcal{S}, \mathcal{A}, \mathcal{O}, T\rangle$, where $\mathcal{A}$ is the action space, $\mathcal{O}$ is the egocentric observation space, and $T:\mathcal{S}\times\mathcal{A}\rightarrow\mathcal{S}$ governs scene transitions. Given $(\mathcal{S},p_0,q)$, the agent receives observation $o_t\in\mathcal{O}$ at each timestep, issues action $a_t\in\mathcal{A}$, and induces a trajectory $\tau=(o_0,a_0,o_1,a_1,\ldots)$ until it commits to a final answer $\hat{y}$ within a budget of $T_{\max}=30$ steps; we validate this budget in Appendix~\ref{app:step_budget_ablation}. The action space spans perception, locomotion, and manipulation. Agents may move through the scene, rotate their viewpoint, interact with objects, and terminate with \texttt{answer($\hat{y}$, $c$)}, which commits to an answer $\hat{y}$ with confidence $c$. The full action space vocabulary is in Figure~\ref{fig:action_space}; Figure~\ref{fig:dataset_example} shows an example trajectory; and Appendix~\ref{app:high_level_actions} discusses the rationale for high-level action design. Although answers are free-form, the question phrasing implicitly specifies the expected format, such as yes/no for relational tasks, a category for comparisons, an integer for counting, or an ordering for procedural tasks. A response is correct if $\hat{y}=y^*$.

\vspace{-0.5em}
\paragraph{Simulation Environment.}
We build \textsc{ESI-Bench} on BEHAVIOR-1K within the OmniGibson simulator. BEHAVIOR-1K provides 51 interactive 3D scenes spanning residential, commercial, and institutional environments, totaling over 300 rooms and 9k object instances across 1,829 categories, with physical properties such as friction, mass, and articulation. OmniGibson, built on NVIDIA Isaac Sim and PhysX 5, supports embodied spatial evaluation through rigid-body contact physics, particle-based fluids, transparency rendering, realistic lighting and reflections, and extended object states such as fill levels and toggled states. For each task instance, we randomly sample a BEHAVIOR-1K scene and select rooms based on room type and task-category requirements from a combined room-object list. We load the room into OmniGibson, allow physics to settle, and query the simulator state to extract a structured scene graph with object bounding boxes, categories, spatial relationships, room assignments, and states such as fillable capacity, toggled state, and contact flags. This scene graph supports scenario construction by providing the object inventory for task-relevant selection, the spatial layout for computing agent and camera poses, and the geometric and object-state ground truth for deriving labels. Appendix~\ref{app:synthetic_simulator} discusses the rationale for using such simulation environment.


 \vspace{-0.5em}
\subsection{Task Construction}
\vspace{-0.5em}
\paragraph{Task Proposal.}
GPT-4o is prompted with the scene graph alongside task category requirements to select task-relevant objects from a random sample of 200 candidate categories drawn from the full BEHAVIOR-1K inventory, applying task-specific physical criteria to choose the most appropriate categories and resolve a specific model instance per category. Beyond object selection, GPT-4o also determines the initial positions of both the objects and the agent within the scene, and generates a ground-truth action trajectory providing the optimal sequence of actions needed to resolve the task. The selected objects and their spatial configuration implicitly define the task, with the ground-truth answer $y^*$ derived directly from the resulting scene state.

\vspace{-0.7em}
\paragraph{Scene Instantiation.}
Given GPT-4o-proposed object selections and initial positions, we load objects into the scene. Before placement, we check conflicts with existing scene content using bbox intersection tests. Objects are then placed on supporting surfaces via physics-based kinematic sampling and settle for a fixed number of simulation steps. After settling, we validate each configuration through stability checks from re-queried bboxes, per-view object existence checks using segmentation masks, and contact-flag validation when applicable. Configurations failing any check are rejected.

\vspace{-0.7em}
\paragraph{Agent Trajectory Collection.}
The agent is initialized at the GPT-4o-proposed pose, sampled using controlled randomization and predefined placement rules designed to withhold the scene configuration and properties. At initialization, we apply the same verification battery used for scene instantiation: per-view object existence checks, bbox  re-querying, and contact-flag validation when applicable. Proposed actions are executed step by step, with each step rendered as an egocentric observation and verified using the same checks. Trajectories failing verification at any step are discarded.

\vspace{-0.7em}
\paragraph{Metadata Saving.}
Upon completing trajectory collection, we save each task instance to a structured JSON file containing the scene, room, floor, per-object category and model instance, verified initial position and quaternion, agent initial pose and quaternion $p_0$, per-view object existence flags, question, ground-truth answer, and action trajectory. This metadata provides a self-contained, reproducible record of the task instance that can be directly reloaded into  the BEHAVIOR environment.

\vspace{-0.7em}
\paragraph{Human Verification and Generator Bias Audit.}
All generated task instances are reviewed by human annotators using rendered per-step observations and metadata. Annotators verify three criteria: correctness, ensuring the initial state and trajectory are physically valid; answerability, ensuring the task is solvable through interaction and spatially unambiguous; and non-triviality, ensuring the task cannot be solved from visual bias or prior knowledge alone and requires genuine spatial uncertainty. Each instance is independently reviewed by three annotators, with disagreements resolved by majority vote. Instances failing any criterion are discarded. Because GPT-4o is used as a proposal engine for objects, placements, questions, and trajectories, we further audit the generated tasks for possible linguistic and object-category biases. Appendix~\ref{app:human_verification_generator_bias} reports detailed human verification protocol, verification scores, shortcut baselines, diversity statistics, and comparison between GPT and matched human-generated tasks. These results show that GPT-4o-generated tasks are high-quality, exhibit limited shortcut bias, and have similar difficulty to human-generated tasks.

\begin{figure}[t]
\centering
\begin{subfigure}[c]{0.63\textwidth}
\centering
\scriptsize
\setlength{\tabcolsep}{4pt}
\renewcommand{\arraystretch}{0.9}
\begin{tabular}{>{\centering\arraybackslash}p{2.6cm}p{6cm}}
\midrule
\textbf{Category} & \textbf{Definition} \\
\midrule
\cellcolor{orangeBg}\textbf{Physical Structure} & \cellcolor{orangeBg}Reveal hidden containment capacity through manipulation \\
\midrule
\cellcolor{orangeBg}\textbf{Physical Dynamics} & \cellcolor{orangeBg}Predict stability and motion under physical forces \\
\midrule
\cellcolor{orangeBg}\textbf{Specular Reflection} & \cellcolor{orangeBg}Infer object relations from mirror reflections \\
\midrule
\cellcolor{orangeBg}\textbf{Perceptual Grounding} & \cellcolor{orangeBg}Resolve viewpoint-dependent perceptual ambiguities \\
\midrule
\cellcolor{purpleBg}\textbf{Metric Comparison} & \cellcolor{purpleBg}Overcome forced-perspective distortions in size and distance \\
\midrule
\cellcolor{purpleBg}\textbf{Spatial Relations} & \cellcolor{purpleBg}Navigate to vantage points that break projective symmetry \\
\midrule
\cellcolor{purpleBg}\textbf{Cognitive Mapping} & \cellcolor{purpleBg}Construct topological representations through locomotion \\
\midrule
\cellcolor{blueBg}\textbf{Enumerative Perception} & \cellcolor{blueBg}Count objects despite occlusion, merging, or enclosure \\
\midrule
\cellcolor{greenBg}\textbf{Temporal Understanding} & \cellcolor{greenBg}Infer state changes through interaction and observation \\
\midrule
\cellcolor{greenBg}\textbf{Action Sequencing} & \cellcolor{greenBg}Reason over action sequences and causal dependencies \\
\midrule
\end{tabular}
\caption{ESI-Bench task categories and definitions.}
\label{fig:category_table}
\end{subfigure}%
\hfill%
\begin{subfigure}[c]{0.37\textwidth}
\centering
\includegraphics[width=\textwidth]{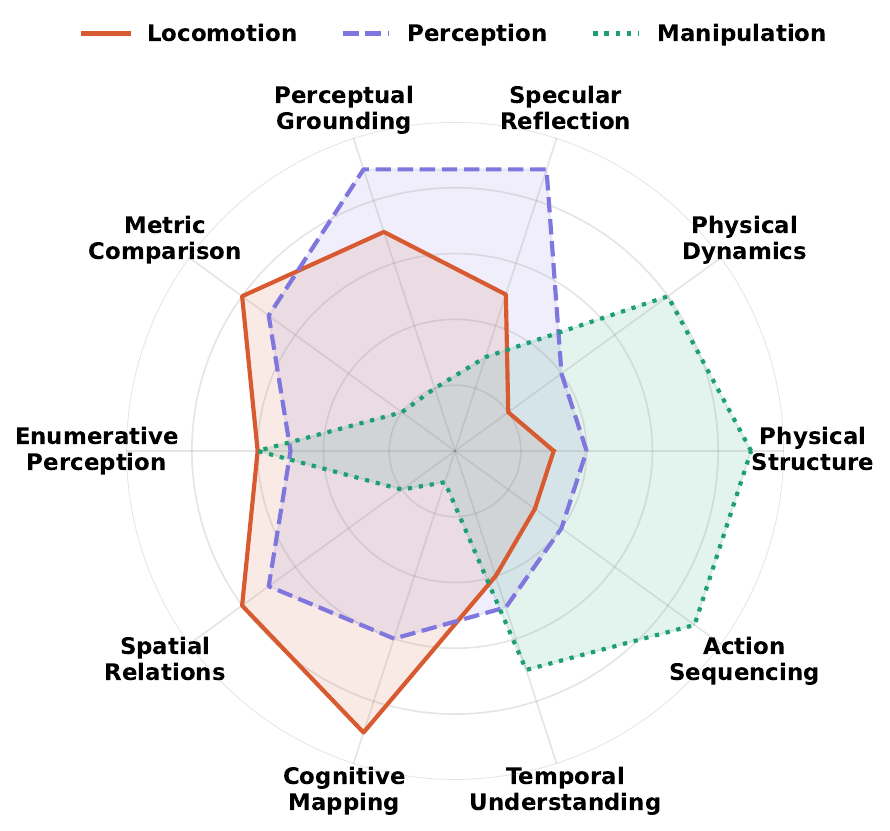}
\caption{Required action types per category.}
\label{fig:radar_chart}
\end{subfigure}
\vspace{-0.5em}
\caption{ESI-Bench task categories (L). Combination and level of embodied action types (R).}
\label{fig:category_overview}
\vspace{-1.5em}
\end{figure}

\vspace{-0.5em}
\subsection{Task Categories and Statistics}
\vspace{-0.5em}
ESI-Bench comprises 3,081 tasks across 10 categories and 29 subcategories. The category definitions are shown in Figure~\ref{fig:category_table}. Due to space constraints, sub-category definitions is provided in Appendix~\ref{sec:task_description}. Figure~\ref{fig:task_distribution} reports the task distribution, and Figure~\ref{fig:teaser} provides concrete examples for each subcategory. 
Per-category subcategory distributions are shown in Appendix~\ref{sec:task_construction_details}. Each category demands a distinct combination of embodied abilities, as illustrated in Figure~\ref{fig:radar_chart}.

\vspace{-0.5em}
\section{Experiments}
\vspace{-0.5em}


\definecolor{orangeBg}{RGB}{255,239,214}
\definecolor{blueBg}{RGB}{222,236,255}
\definecolor{purpleBg}{RGB}{239,224,246}
\definecolor{greenBg}{RGB}{224,244,237}

\newcommand{\nd}{--}
\newcommand{\rotang}{70}

\newcolumntype{C}{>{\centering\arraybackslash}p{0.82cm}}

\newcommand{\mh}[2]{%
  \rotatebox[origin=lb]{\rotang}{%
    \parbox[c][0.72cm][c]{0.42cm}{%
      \centering\footnotesize\itshape #1\\[-1pt]\footnotesize\itshape #2%
    }%
  }%
}

\newcommand{\catrow}[2]{%
  \rowcolor{#1}
  \multicolumn{21}{@{}l@{}}{\footnotesize\bfseries #2}\\[-1pt]
}

\newcommand{\subcat}[1]{\hspace{0.6em}\footnotesize\itshape #1}

\begin{table*}[t]
\centering
\small
\caption{\textbf{\textsc{ESI-Bench} results (\%)} across passive, active, and oracle paradigms for 2D+VLMs, 3D+LLMs, and humans, showing active exploration benefits and the gap to oracle action selection.}
\vspace{-0.4em}
\label{tab:esi}

\resizebox{\textwidth}{!}{%
\setlength{\tabcolsep}{0pt}
\renewcommand{\arraystretch}{1.18}

\begin{tabular}{@{}p{3.3cm}*{20}{C}@{}}
\toprule

& \multicolumn{8}{c}{\textbf{2D + VLM}}
& \multicolumn{8}{c}{\textbf{3D + LLM}}
& \multicolumn{4}{c}{\textbf{Human Performance}} \\

\cmidrule(lr){2-9}\cmidrule(lr){10-17}\cmidrule(lr){18-21}

& \multicolumn{4}{c}{\textbf{GPT-5}}
& \multicolumn{4}{c}{\textbf{Gemini 3.1}}
& \multicolumn{4}{c}{\textbf{VGGT + Gemini}}
& \multicolumn{4}{c}{\textbf{GT 3D + Gemini}}
& \multicolumn{4}{c}{} \\

\cmidrule(lr){2-5}\cmidrule(lr){6-9}
\cmidrule(lr){10-13}\cmidrule(lr){14-17}

\rule{0pt}{3.0em}\footnotesize\textbf{Category / Subcategory}
& \mh{Passive}{Single} & \mh{Passive}{Multiple} & \mh{Active}{Multiple} & \mh{GT}{Passive}
& \mh{Passive}{Single} & \mh{Passive}{Multiple} & \mh{Active}{Multiple} & \mh{GT}{Passive}
& \mh{Passive}{Single} & \mh{Passive}{Multiple} & \mh{Active}{Multiple} & \mh{GT}{Passive}
& \mh{Passive}{Single} & \mh{Passive}{Multiple} & \mh{Active}{Multiple} & \mh{GT}{Passive}
& \mh{Passive}{Single} & \mh{Passive}{Multiple} & \mh{Active}{Multiple} & \mh{GT}{Passive} \\

\midrule


\catrow{orangeBg}{Perceptual Grounding}
\rowcolor{orangeBg}
\subcat{View Hallucination}
& 11.7 & 20.2 & 60.1 & 87.8
& 39.9 & 32.9 & 68.1 & 91.1
& 56.3 & 59.0 & 76.1 & 87.6
& 61.6 & 60.6 & 74.1 & 89.0
&  52.2 & 45.8 & 80.8 & 83.8 \\
\rowcolor{orangeBg}
\subcat{Partial Occlusion}
& 32.6 & 26.3 & 47.4 & 86.3
& 30.5 & 32.6 & 70.5 & 88.4
& 51.6 & 48.4 & 53.7 & 86.3
& 56.8 & 55.8 & 72.6 & 91.6
& 37.9 & 34.7 & 80.0 & 87.4 \\
\rowcolor{orangeBg}
\subcat{Material Transparency}
& 30.3 & 36.7 & 66.1 & 96.3
& 44.0 & 45.0 & 52.3 & 88.0
& 37.4 & 29.4 & 31.8 & 90.9
& 27.8 & 31.2 & 60.4 & 100
& 41.3 & 44.0 & 93.6 & 97.2 \\

\midrule

\catrow{orangeBg}{Physical Structure}
\rowcolor{orangeBg}
\subcat{Rigid Containment}
& 45.0 & 42.5 & 42.5 & 95.0
& 47.5 & 40.0 & 67.5 & 97.5
& 27.5 & 37.5 & 57.5 & 72.5
& 45.0 & 42.5 & 65.0 & 95.0
& 47.5 & 42.5 & 92.5 & 100 \\
\rowcolor{orangeBg}
\subcat{Liquid Volume}
& 66.2 & 66.2 & 81.6 & 86.0
& 69.9 & 67.6 & 80.9 & 86.8
& 65.4 & 56.7 & 71.3 & 77.2
& 74.5 & 74.5 & 83.1 & 86.8
& 79.4 & 66.9 & 89.7 & 86.8 \\
\rowcolor{orangeBg}
\subcat{Deformable}
& 42.9 & 41.8 & 55.1 & 75.5
& 34.7 & 41.8 & 43.9 & 78.6
& 43.9 & 42.9 & 49.0 & 81.6
& 46.9 & 54.1 & 99.0 & 99.0
& 57.1 & 60.2 & 76.5 & 82.7 \\

\midrule

\catrow{orangeBg}{Physical Dynamics}
\rowcolor{orangeBg}
\subcat{Inclined Plane}
& 57.4 & 60.7 & 77.0 & 86.9
& 65.6 & 62.3 & 83.6 & 88.5
& 67.2 & 62.3 & 80.3 & 86.9
& 63.9 & 63.9 & 83.6 & 91.8
& 60.7 & 62.3 & 83.6 & 83.6\\
\rowcolor{orangeBg}
\subcat{Stacking \& Stability}
& 34.8 & 37.1 & 62.9 & 86.5
& 38.2 & 36.0 & 78.7 & 84.3
& 34.8 & 39.3 & 55.1 & 62.9
& 27.2 & 33.7 & 68.5 & 86.5
& 36.0 & 39.3 & 84.3 & 86.5 \\

\midrule

\catrow{orangeBg}{Specular Reflection}
\rowcolor{orangeBg}
\subcat{Reflection Authoring}
& 68.7 & 70.7 & 70.3 & 73.7
& 60.6 & 60.6 & 64.9 & 67.0
& 52.5 & 54.6 & 53.5 & 55.6
& 55.6 & 58.6 & 55.6 & 60.6
& 94.9 & 88.9 & 96.0 & 96.0 \\
\rowcolor{orangeBg}
\subcat{Spatial Relations}
& 50.4 & 38.9 & 54.8 & 58.4
& 43.4 & 42.5 & 44.2 & 46.9
& 41.6 & 39.8 & 41.6 & 43.4
& 38.9 & 35.4 & 54.2 & 41.6
& 78.8 & 78.8 & 84.1 & 87.6 \\
\rowcolor{orangeBg}
\subcat{Correspondence}
& 39.8 & 51.1 & 52.3 & 56.8
& 37.5 & 40.9 & 42.0 & 42.0
& 37.5 & 43.2 & 48.9 & 48.9
& 31.8 & 39.8 & 42.3 & 48.9
& 85.2 & 80.7 & 89.8 & 92.0 \\

\midrule


\catrow{purpleBg}{Spatial Relations}
\rowcolor{purpleBg}
\subcat{Linear Alignment}
& 27.7 & 31.9 & 42.6 & 60.6
& 47.9 & 44.6 & 67.0 & 77.7
& 45.7 & 38.3 & 53.2 & 59.6
& 73.4 & 73.4 & 84.0 & 89.4
& 51.1 & 39.4 & 73.4 & 79.8 \\
\rowcolor{purpleBg}
\subcat{Geometric Configuration}
& 25.3 & 20.4 & 26.0 & 26.0
& 27.5 & 22.3 & 27.5 & 44.6
& 9.9 & 11.3 & 18.6 & 32.6
& 70.8 & 73.6 & 88.0 & 100
& 32.4 & 33.5 & 74.3 & 86.3 \\
\rowcolor{purpleBg}
\subcat{Physical Contact}
& 40.0 & 41.7 & 64.2 & 90.0
& 60.8 & 55.8 & 70.0 & 74.2
& 59.2 & 41.7 & 59.2 & 78.2
& 65.8 & 67.5 & 70.8 & 72.5
& 35.8 & 40.8 & 88.3 & 90.8 \\

\midrule

\catrow{purpleBg}{Metric Comparison}
\rowcolor{purpleBg}
\subcat{Dimensional Size}
& 42.5 & 44.9 & 67.7 & 80.3
& 44.3 & 41.3 & 68.3 & 82.6
& 50.4 & 40.1 & 59.3 & 80.8
& 58.6 & 60.5 & 69.9 & 85.9
& 48.5 & 56.9 & 82.6 & 91.9 \\
\rowcolor{purpleBg}
\subcat{Spatial Distance}
& 53.9 & 49.1 & 58.6 & 73.7
& 52.6 & 50.5 & 59.9 & 80.3
& 50.7 & 47.3 & 55.0 & 71.4
& 59.6 & 61.2 & 64.5 & 96.7
& 57.9 & 59.9 & 69.1 & 78.9 \\

\midrule

\catrow{purpleBg}{Cognitive Mapping}
\rowcolor{purpleBg}
\subcat{Connectivity}
& 68.3 & 70.0 & 68.3 & 78.3
& 51.7 & 48.3 & 60.0 & 85.0
& 66.7 & 66.7 & 73.3 & 83.3
& 65.0 & 65.0 & 71.7 & 86.7
& 68.3 & 68.3 & 81.7 & 91.7 \\
\rowcolor{purpleBg}
\subcat{Traversable Passage}
& 68.3 & 66.7 & 71.7 & 73.3
& 66.7 & 61.7 & 73.3 & 78.3
& 68.3 & 63.3 & 66.7 & 78.3
& 71.7 & 73.3 & 71.7 & 80.0
& 70.0 & 71.7 & 78.3 & 85.0 \\
\rowcolor{purpleBg}
\subcat{Regional Boundary}
& 65.0 & 63.8 & 65.0 & 67.5
& 63.8 & 62.5 & 65.0 & 65.0
& 65.0 & 65.0 & 62.5 & 67.5
& 60.0 & 68.8 & 62.5 & 68.8
& 61.3 & 62.5 & 67.5 & 70.0 \\
\rowcolor{purpleBg}
\subcat{Long-Term Navigation}
& 40.0 & 38.3 & 41.7 & 50.0
& 33.3 & 35.0 & 33.3 & 41.7
& 36.7 & 40.0 & 43.3 & 51.6
& 36.7 & 41.7 & 43.3 & 40.0
& 35.0 & 36.7 & 51.7 & 61.7 \\

\midrule


\catrow{blueBg}{Enumerative Perception}
\rowcolor{blueBg}
\subcat{Counting w Occlusion}
& 3.3 & 3.3 & 13.3 & 56.7
& 3.3 & 3.3 & 10.0 & 63.3
& 0.0 & 0.0 & 13.3 & 53.3
& 33.3 & 80.0 & 83.3 & 100.0
& 6.7 & 10.0 & 53.3 & 76.7 \\
\rowcolor{blueBg}
\subcat{Spatial Segmentation}
& 3.3 & 6.7 & 26.7 & 63.3
& 3.3 & 3.3 & 16.7 & 70.0
& 3.3 & 3.3 & 23.3 & 56.7
& 43.3 & 80.0 & 66.7 & 86.7
& 13.3 & 23.3 & 63.3 & 70.0 \\
\rowcolor{blueBg}
\subcat{Merged Observation}
& 38.3 & 35.0 & 51.7 & 50.0
& 43.3 & 51.7 & 61.7 & 75.0
& 63.3 & 51.7 & 55.0 & 65.0
& 51.7 & 63.3 & 93.3 & 100.0
& 60.0 & 60.0 & 65.0 & 73.3 \\
\rowcolor{blueBg}
\subcat{Category Ambiguity}
& 8.3 & 10.0 & 8.3 & 41.7
& 13.3 & 13.3 & 15.0 & 48.3
& 21.7 & 26.7 & 26.7 & 46.7
& 18.3 & 76.7 & 40.0 & 93.3
& 23.3 & 25.0 & 51.7 & 61.7 \\
\rowcolor{blueBg}
\subcat{Structural Enclosure}
& 5.0 & 10.0 & 22.5 & 67.5
& 2.5 & 0.0 & 10.0 & 52.5
& 0.0 & 0.0 & 12.5 & 52.5
& 40.0 & 60.0 & 50.0 & 100.0
& 10.0 & 22.5 & 42.5 & 77.5 \\
\rowcolor{blueBg}
\subcat{Illumination Variability}
& 6.0 & 22.0 & 22.0 & 46.0
& 12.0 & 16.0 & 22.0 & 58.0
& 10.0 & 28.0 & 30.0 & 40.0
& 20.0 & 84.0 & 42.0 & 96.0
& 20.0 & 34.0 & 62.0 & 66.0\\

\midrule


\catrow{greenBg}{Temporal Understanding}
\rowcolor{greenBg}
\subcat{Unobserved Change}
& 40.5 & 41.2 & 51.4 & 77.0
& 37.2 & 37.8 & 47.3 & 74.5
& 37.8 & 41.9 & 45.8 & 76.6
& 39.9 & 42.6 & 46.5 & 74.5
& 39.4 & 39.4 & 70.8 & 81.0 \\
\rowcolor{greenBg}
\subcat{Agent Observation}
& 40.6 & 30.1 & 51.0 & 72.7
& 37.6 & 36.1 & 58.6 & 65.4
& 32.7 & 27.9 & 34.2 & 56.5
& 36.3 & 33.3 & 76.7 & 87.8
& 38.3 & 42.1 & 83.5 &  90.2 \\

\midrule

\catrow{greenBg}{Action Sequencing}
\rowcolor{greenBg}
\subcat{Action Order Inference}
& 36.4 & 37.7 & 41.6 & 67.5
& 44.2 & 46.8 & 54.5 & 72.7
& 35.1 & 31.1 & 50.6 & 58.4
& 44.2 & 46.8 & 51.9 & 75.3
& 40.3 & 41.6 & 74.0 & 81.8 \\

\bottomrule
\end{tabular}}
\vspace{-2.2em}
\end{table*}

\subsection{Models and Evaluation Setup}
\vspace{-0.5em}

We evaluate models across four paradigms, organized by the degree of action and perceptual access granted to the agent: (1) \textbf{Passive Single-View} provides a single fixed observation from the initial pose, establishing a baseline under conditions identical to existing spatial benchmarks; (2) \textbf{Passive Multi-View} provides a set of 30 views along a randomly-sampled trajectory starting at the initial pose in the scene, matching the maximum active trajectory budget $T_{\max}=30$, to broadly cover the full environment and simulate exhaustive passive scene coverage without any agent active action or viewpoint selection; (3) \textbf{Active Exploration} places the agent the initial pose with full access to the action space, requiring it to gather evidence through deliberate movement and interaction before committing to a final answer;  (4) \textbf{Ground-Truth Passive} provides the sequence of views rendered along the ground-truth action trajectory, serving as an oracle ablation that \textit{separates perception errors from action errors}. Comparing (1) vs. (2) reveals whether more passive views helps;  comparing (2) vs. (3) isolates the benefit of active action-guided over passive exhaustive coverage; comparing (3) vs. (4) isolates whether failures stem from the agent's inability to select informative actions or from perceptual limitations given the perfect views themselves.

\begin{figure}[t]
    \centering
    \includegraphics[width=\textwidth]{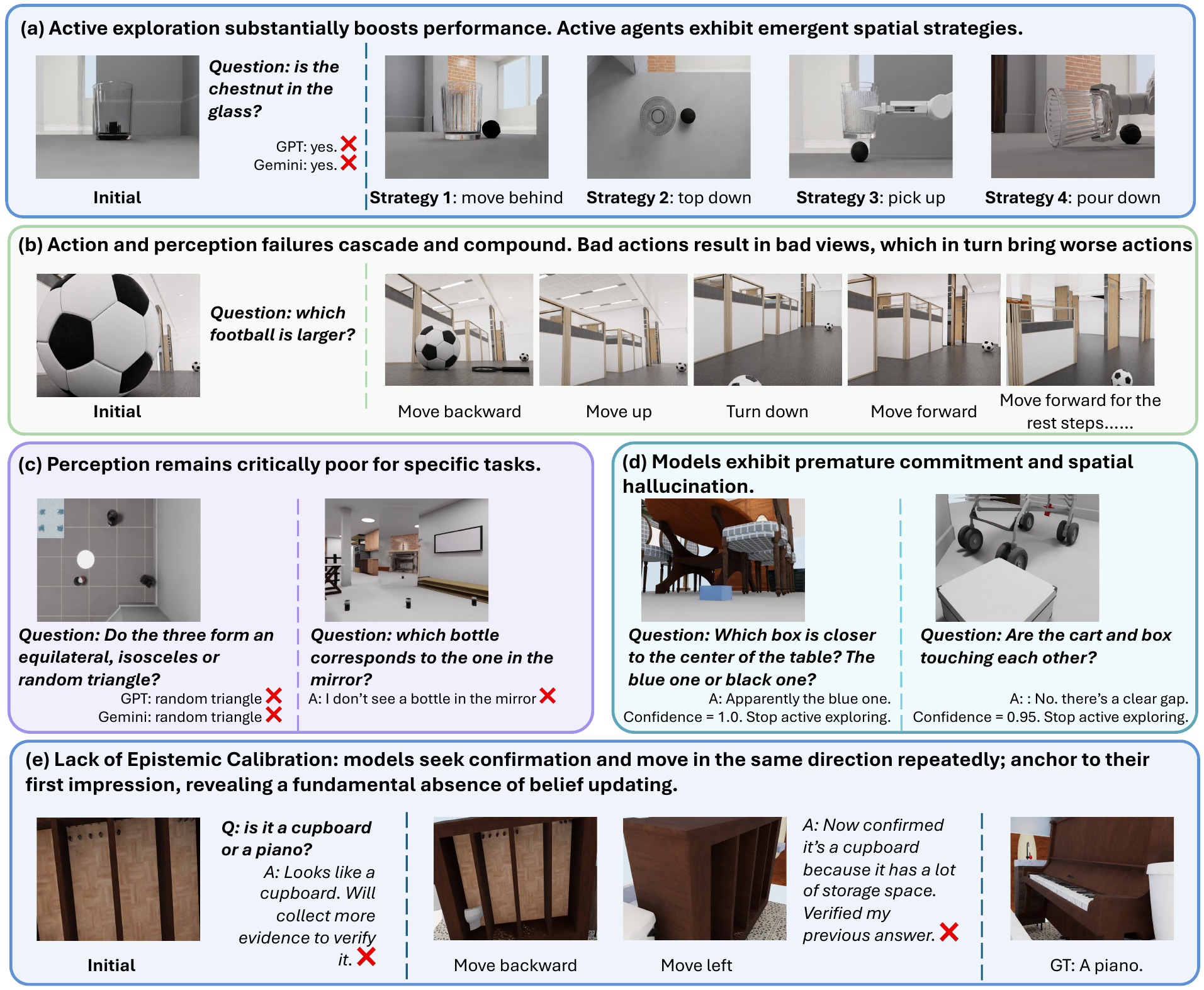}
    \caption{\textbf{Qualitative Study}, showing success \& failure modes and reasons behind model behavior.}
    \label{fig:esi_qualitative}
    \vspace{-1.7em}
\end{figure}

We evaluate two families of models. 2D vision-language models include GPT-5 and Gemini 3.1, each taking egocentric visual observations as input, evaluated across all four paradigms. 3D-augmented models include VGGT+Gemini, where explicit 3D scene representations are reconstructed from multi-view observations via VGGT, from which scene graphs are constructed and provided to the language model; and Ground-Truth 3D+Gemini, where perfect point clouds derived directly from simulator state are used to construct scene graphs provided to the language model, serving as an oracle ablation for 3D grounding. Human performance is additionally collected on all tasks to establish a human upper bound. The full human study protocol is described in Appendix~\ref{appendix:human_study}. All models are evaluated zero-shot with no task-specific fine-tuning, and results are reported in Table~\ref{tab:esi} as accuracy on matched subsets to ensure fair cross-paradigm comparison, with qualitative examples in Figure~\ref{fig:esi_qualitative} and Appendix~\ref{more_qualitative}. More model details including prompting templates are in Appendix~\ref{app:active_agent_prompting}.

\newcommand{\findingblock}[2]{
\begin{tcolorbox}[
  enhanced,
  colback=blue!3!white,
  colframe=blue!20!gray!40!white,
  borderline west={4pt}{0pt}{blue!30!gray!90!black},
  boxrule=0.3pt,
  arc=4pt,
  left=8pt, right=8pt, top=6pt, bottom=3pt,
  before skip=8pt, after skip=4pt,
  title={\faLightbulb[regular]\enspace\textbf{#1}},
  fonttitle=\small\bfseries\color{white},
  colbacktitle=blue!30!gray!90!black,
  attach boxed title to top left={yshift=-2mm, xshift=6pt},
  boxed title style={arc=3pt, boxrule=0pt}
]
\begin{itemize}[leftmargin=12pt, itemsep=2pt, topsep=0pt, parsep=0pt]
#2
\end{itemize}
\end{tcolorbox}
}

\subsection{What Really Holds Spatial Intelligence Back: Seeing or Acting?}
\vspace{-0.5em}

\begin{figure}[t]
    \centering
    \includegraphics[width=0.9\linewidth]{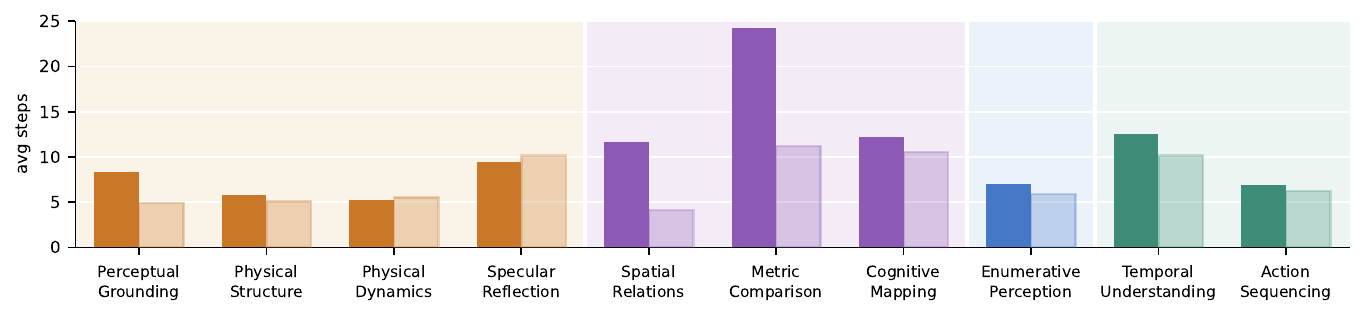}
    \vspace{-0.5em}
    \caption{Average number of active exploration steps to reach a correct answer for GPT-5 (solid) and Gemini 3.1 (outlined), grouped by Spelke's four core knowledge systems.}
    \vspace{-1.5em}
    \label{fig:avg_steps}
\end{figure}

Table~\ref{tab:esi} reports accuracy across GPT-5 and Gemini 3.1 under passive and active evaluation paradigms. Without any explicit instruction, active agents \textit{spontaneously discover emergent spatial strategies}: to determine whether a chestnut is inside a glass, agents independently develop four distinct approaches, moving behind the object, repositioning top-down, picking it up, and pouring it out (Figure~\ref{fig:esi_qualitative}a), none of which were prescribed. These emergent abilities drive consistent and substantial gains across tasks: 
on View Hallucination, from 39.9\% to 68.1\%;
and on Rigid Containment, from 47.5\% to 67.5\%. In contrast, passive multi-view provides negligible or negative gains despite consuming far more images (\textit{e.g.}, GPT-5 drops from 53.9\% to 49.1\% on Spatial Distance compared with the single-view passive baseline), confirming that observation quantity without selective action often adds noise rather than signal. Figure~\ref{fig:avg_steps} further supports this distinction: active exploration reaches correct answers with only a small number of task-conditioned steps, suggesting that the benefit comes from selective evidence acquisition rather than simply increasing the number of observations. We quantify the diversity of emergent strategies in Appendix~\ref{sec:emergent_behaviors} and analyze the sources of passive multi-view degradation in Appendix~\ref{app:passive_multiview_analysis}.

Ground-truth passive trajectories reveal that the bottleneck is largely in action selection, not perception: GPT-5 improves from 42.5\% under active exploration to 95.0\% with ground-truth trajectory views on Rigid Containment,
and from 64.2\% to 90.0\% on Physical Contact. Geometric Configuration and Specular Reflection stand as exceptions where oracle views provide limited benefit: GPT-5 reaches only 26.0\% on Geometric Configuration under ground-truth passive, and models consistently fail to determine whether three objects form an equilateral triangle even from the perfect viewpoint (Figure~\ref{fig:esi_qualitative}c); on Specular Reflection, models hallucinate objects in mirrors that do not exist, or fail to identify the correct real-world correspondence altogether (Figure~\ref{fig:esi_qualitative}c). These cases indicate hard perceptual limits that no action strategy can overcome. The active-to-oracle gap further shows that action and perception failures cascade and compound: on Counting w Occlusion, the GPT-5 gap reaches 43.4 points, and on Structural Enclosure, 45.0 points. Bad actions lead to bad views, which in turn induce worse subsequent actions that cannot be recovered within the step budget (Figure~\ref{fig:esi_qualitative}b).

\findingblock{Action Blindness Dominates Perceptual Blindness, while Their Coupling Drives Failure Cascades}{
  \item Without explicit instruction, agents spontaneously acquire spatial strategies, yielding substantial gains over passive multi-view, which often adds noise rather than signal.
  \item For most tasks, perception is not the bottleneck: with the right viewpoint, the agent succeeds; yet some tasks hit a hard perceptual ceiling that no action can overcome.
  \item Suboptimal actions produce bad views, which in turn produce worse actions, cascading into reasoning failures unrecoverable within the step budget.
}

\subsection{When Does 3D Help, and When Does It Hurt?}
\vspace{-0.5em}
From Table~\ref{tab:esi}, ground-Truth 3D+Gemini substantially improves performance when the underlying 3D representation is accurate: on Geometric Configuration, it reaches 70.8\% compared with 27.5\% for Gemini 3.1, a 43.3-point gain; on Counting w Occlusion, it reaches 33.3\% compared with 3.3\%, a 30.0-point gain. These gains suggest that explicit geometry can resolve spatial ambiguities that are difficult to infer from 2D observations alone. However, imperfect 3D reconstruction can be actively harmful: VGGT+Gemini drops to 9.9\% on Geometric Configuration compared with Gemini 3.1's 27.5\%, and to 0.0\% on Counting w Occlusion compared with 3.3\%. Rather than merely failing to help, noisy reconstructions distort fine-grained spatial relations and cause the LLM to reason over a corrupted scene graph, making 3D augmentation a high-variance strategy that amplifies both successes and failures. 
Common failure modes of VGGT-based method are detailed in Appendix~\ref{app:vggt_failure_analysis}.

\findingblock{3D Helps When Geometry Is Perfect, But Imperfect Reconstruction Actively Misleads}{
  \item Explicit 3D grounding improves performance on tasks where depth and occlusion render 2D projections fundamentally ambiguous.
  \item Imperfect 3D reconstruction degrades performance, proving more harmful than 2D baselines.
}

\subsection{How Far Are Models From Human-Level Spatial Reasoning?}
\vspace{-0.5em}
Table~\ref{tab:esi} shows that human and model performance are often comparable when agents are given fixed observations, either from the initial passive view or from the ground-truth trajectory. This suggests that the perceptual gap between humans and models, while present, may be smaller than commonly assumed. In some categories, models even match or exceed human passive performance. Under ground-truth trajectories, Gemini reaches 88.4\% on Partial Occlusion compared with human score of 87.4\%, and GPT-5 reaches 96.3\% on Material Transparency compared with 97.2\% for humans. Yet the divergence emerges sharply under active exploration: humans dramatically outperform models by knowing what to look for and when to stop, with active performance much closer to ground-truth passive performance (\textit{e.g.}, on Physical Contact, humans reach 88.3\% versus 64.2\% for GPT-5; on Material Transparency, humans reach 93.6\% versus 52.3\% for Gemini 3.1). Examining model and human trajectories, we find humans exhibit stronger epistemic caution: they gather more observations before committing, actively seek viewpoints that could falsify their current hypothesis, and reduce confidence under ambiguity. Figure~\ref{fig:esi_qualitative}d shows models terminate exploration prematurely, committing with high confidence after only a few steps regardless of evidence ambiguity, producing spatial hallucinations that contradict the scene state. This overconfidence is compounded by a directional bias in action selection: rather than probing orthogonal angles or seeking views that could disprove an initial impression, models repeatedly move in the same direction, accumulating redundant rather than informative observations. As shown in Figure~\ref{fig:esi_qualitative}e, when asked whether an object is a cupboard or a piano, the model initially guesses cupboard, then moves backward and left only to confirm that belief rather than challenge it, ultimately asserting the wrong answer with increased confidence despite never seeking a falsifying viewpoint. Models anchor rigidly to their first impression rather than revising, exposing a failure of epistemic calibration that is fundamentally distinct from perceptual ability and cannot be resolved by stronger visual encoders or additional exploration steps alone. To make this model-human gap more concrete, Appendix~\ref{app:metacognitive_gap} operationalizes it through trajectory-level quantitative measures of view diversity, contrastive-view seeking, and belief revision.

\findingblock{Lack of Epistemic Calibration: Models Can See But Do Not Know When They Have Seen Enough}{
  \item Models commit prematurely with high confidence; humans treat uncertainty as a signal to keep looking, not a reason to answer.
  \item Humans seek viewpoints that falsify their hypothesis; models seek confirmation and move in the same direction repeatedly.
  \item Humans revise beliefs when contradicted; models anchor to their first impression, revealing a fundamental absence of belief updating.
}

\section{Conclusion}
\vspace{-0.5em}
We introduced \textsc{ESI-Bench}, a benchmark requiring agents to acquire evidence through perception, locomotion, and manipulation. Across 10 categories and 29 subcategories, active exploration improves over passive observation, but models still struggle with action selection, belief revision, and confidence calibration. \textsc{ESI-Bench} provides a testbed for studying agents that close the perception-action loop.

\section*{Acknowledgement}
This paper is funded by ONR MURI N00014-24-1-2748,  also supported by IITP funded by the Korean Government (MSIT) (No. RS-2024-00457882, National AI Research Lab Project), and Grant AW1134392 (Reasoning in Motion) from the TRI University 3.0 Program. This work used computational resources provided by Google through the Google Gemini Academic Program.

\bibliographystyle{plainnat}   
\bibliography{references}      

\appendix
\clearpage

{
    \hypersetup{linkcolor=black}
    \tableofcontents
}
\clearpage

\section{Contribution Statement}
Yining Hong is responsible for coming up with the idea; code writing for Perceptual Grounding, Physical Structure (Rigid + Liquid), Physical Dynamics, Spatial Relations, Metric Comparison, Agent Observation and Action Sequencing; the full paper writing and figure drawing; code structuring; coordinating among authors; website etc. Jiageng Liu wrote the codes for Deformable, Specular Reflection, Cognitive Mapping, Enumerative Perception, Unobserved Change; generated the croissant file; code cleaning; did the demo video etc. Han Yin wrote the related works section, drew some of the original figures which were then modified by Yining. The other people took advising roles. 

\section{Broader Impacts \& Limitations}
\label{appendix:broader_impacts}

\textsc{ESI-Bench} advances the evaluation of embodied spatial intelligence by requiring agents to close the perception-action loop rather than reason from pre-given observations. The primary societal benefit is diagnostic: by exposing systematic failures in epistemic calibration and action selection, the benchmark provides concrete targets for improving the safety and reliability of embodied systems deployed in homes, hospitals, warehouses, and assistive robotics. Agents that commit prematurely with high confidence despite insufficient evidence pose real risks in physical deployment; \textsc{ESI-Bench} makes this failure mode measurable and therefore addressable.
The benchmark itself poses minimal misuse risk. All scenes are synthetically generated within OmniGibson using the BEHAVIOR-1K scene pool; no real-world personal data, biometric information, or scraped internet content is involved. The task design targets spatial reasoning competence rather than any sensitive domain, and the evaluation is zero-shot with no fine-tuning that could encode harmful behaviors.
One limitation is rendering realism. Although OmniGibson provides realistic scenes and physics, visual appearance still differs from real camera data in texture quality, lighting, motion blur, and sensor artifacts.

\section{Per-Category Task Description}
In Table \ref{sec:task_description}, we show the detailed definitions of each sub-category in our \textsc{ESI-Bench}.

\label{sec:task_description}
\begin{table*}[htbp]
\centering
\tiny
\caption{\textbf{\textsc{ESI-Bench} task taxonomy and definitions} of 10 categories spanning 29 subcategories. Colors correspond to Spelke's four core knowledge systems: \textcolor{esiOrange}{object representation}, \textcolor{esiPurple}{layout and geometry}, \textcolor{esiBlue}{number representation}, and \textcolor{esiGreen}{agents and goal-directed actions}.}
\label{tab:taxonomy}
\setlength{\tabcolsep}{3pt}
\begin{tabular}{>{\centering\arraybackslash}p{1.65cm}p{2.1cm}p{5.7cm}p{3.8cm}}
\toprule
\textbf{Category} & \textbf{Subcategory} & \textbf{Description} & \textbf{Example} \\
\midrule

\cellcolor{orangeBg}\textcolor{esiOrange}{\textbf{Physical}} & \cellcolor{orangeBg}Rigid Containment & \cellcolor{orangeBg}Plan the placement of multiple objects across multiple containers & \cellcolor{orangeBg}How can all the toys be fit into the boxes? \\
\cellcolor{orangeBg}\textcolor{esiOrange}{\textbf{Structure}} & \cellcolor{orangeBg}Liquid Volume & \cellcolor{orangeBg}Compare liquid-holding capacity of containers & \cellcolor{orangeBg}Which container has larger volume? \\
\cellcolor{orangeBg} & \cellcolor{orangeBg}Deformable  & \cellcolor{orangeBg}Infer structure from deformable object & \cellcolor{orangeBg}What's under the sweater, apple or mug? \\
\midrule

\cellcolor{orangeBg}\textcolor{esiOrange}{\textbf{Physical}} & \cellcolor{orangeBg}Inclined Plane & \cellcolor{orangeBg}Predict object motion and stability on slopes & \cellcolor{orangeBg}Can the apple sit stable on the slope? \\
\cellcolor{orangeBg}\textcolor{esiOrange}{\textbf{Dynamics}} & \cellcolor{orangeBg}Stacking \& Stability & \cellcolor{orangeBg}Whether objects stack or balance given shape, mass and geometry & \cellcolor{orangeBg}How to stack and stabilize these three objects? \\
\midrule

\cellcolor{orangeBg}\textcolor{esiOrange}{\textbf{Specular}} & \cellcolor{orangeBg}Reflection Authoring & \cellcolor{orangeBg}Distinguish real objects from mirror reflections & \cellcolor{orangeBg}Is the object real or reflected? \\
\cellcolor{orangeBg}\textcolor{esiOrange}{\textbf{Reflection}} & \cellcolor{orangeBg}Spatial Relations & \cellcolor{orangeBg}Infer relations among objects across mirror and real-world views & \cellcolor{orangeBg}What relations does the mirror reveal? \\
\cellcolor{orangeBg} & \cellcolor{orangeBg}Correspondence & \cellcolor{orangeBg}Identify which objects appear in mirror given the real-world scene & \cellcolor{orangeBg}Which of the 3 snacks appears in the mirror? \\
\midrule

\cellcolor{orangeBg}\textcolor{esiOrange}{\textbf{Perceptual}} & \cellcolor{orangeBg}Partial Occlusion & \cellcolor{orangeBg}Reason about objects hidden behind other scene elements & \cellcolor{orangeBg}Golf Stick or Umbrella behind the Wall? \\
\cellcolor{orangeBg}\textcolor{esiOrange}{\textbf{Grounding}} & \cellcolor{orangeBg}View Hallucination & \cellcolor{orangeBg}Detect objects whose visibility changes critically with viewing angle & \cellcolor{orangeBg}Cabinet or piano from this view? \\
\cellcolor{orangeBg} & \cellcolor{orangeBg}Material Transparency & \cellcolor{orangeBg}Reason about objects seen through transparent surfaces & \cellcolor{orangeBg}Is the object inside the glass or not? \\
\midrule

\cellcolor{purpleBg}\textcolor{esiPurple}{\textbf{Metric}} & \cellcolor{purpleBg}Dimensional Size & \cellcolor{purpleBg}Compare relative sizes of objects & \cellcolor{purpleBg}Which vase is larger? \\
\cellcolor{purpleBg}\textcolor{esiPurple}{\textbf{Comparison}} & \cellcolor{purpleBg}Spatial Distance & \cellcolor{purpleBg}Compare relative distances with respect to a reference object & \cellcolor{purpleBg}Which flower is closer to the table? \\
\midrule

\cellcolor{purpleBg}\textcolor{esiPurple}{\textbf{Spatial}} & \cellcolor{purpleBg}Linear Alignment & \cellcolor{purpleBg}Whether objects are arranged along a common axis & \cellcolor{purpleBg}Do the glasses form a line? \\
\cellcolor{purpleBg}\textcolor{esiPurple}{\textbf{Relations}} & \cellcolor{purpleBg}Geometric Configuration & \cellcolor{purpleBg}Identify the shape formed by a set of objects & \cellcolor{purpleBg}Do the cups form an equilateral triangle? \\
\cellcolor{purpleBg} & \cellcolor{purpleBg}Physical Contact & \cellcolor{purpleBg}Detect whether two or more objects are in direct contact & \cellcolor{purpleBg}Are the teddy bears touching each other? \\
\midrule

\cellcolor{purpleBg}\textcolor{esiPurple}{\textbf{Cognitive}} & \cellcolor{purpleBg}Connectivity & \cellcolor{purpleBg}Whether two locations or regions are mutually reachable & \cellcolor{purpleBg}Is region A connected to C? \\
\cellcolor{purpleBg}\textcolor{esiPurple}{\textbf{Mapping}} & \cellcolor{purpleBg}Traversable Passage & \cellcolor{purpleBg}Identify navigable corridors or passageways between regions & \cellcolor{purpleBg}Passage between the rooms? \\
\cellcolor{purpleBg} & \cellcolor{purpleBg}Regional Boundary & \cellcolor{purpleBg}Identify and delineate distinct functional spatial regions & \cellcolor{purpleBg}Boundaries of this region? \\
\cellcolor{purpleBg} & \cellcolor{purpleBg}Long-Term Navigation & \cellcolor{purpleBg}Plan multi-step navigation trajectories toward a distant goal & \cellcolor{purpleBg}What room is behind the wall? \\
\midrule

\cellcolor{blueBg}\textcolor{esiBlue}{\textbf{Enumerative}} & \cellcolor{blueBg}Counting w Occlusion & \cellcolor{blueBg}Count objects partially obscured by other scene elements & \cellcolor{blueBg}How many balls are under the blanket? \\
\cellcolor{blueBg}\textcolor{esiBlue}{\textbf{Perception}} & \cellcolor{blueBg}Spatial Segmentation & \cellcolor{blueBg}Count objects separated across distinct spatial regions & \cellcolor{blueBg}One or two cylinders separated by the post? \\
\cellcolor{blueBg} & \cellcolor{blueBg}Category Ambiguity & \cellcolor{blueBg}Count visually similar objects requiring fine-grained distinction & \cellcolor{blueBg}How many apples among the balls? \\
\cellcolor{blueBg} & \cellcolor{blueBg}Merged Observation & \cellcolor{blueBg}Count groups that appear visually merged from a single viewpoint & \cellcolor{blueBg}How many separate stacks of books? \\
\cellcolor{blueBg} & \cellcolor{blueBg}Illumination Variability & \cellcolor{blueBg}Count objects under challenging or non-uniform lighting & \cellcolor{blueBg}Objects in the dim scene? \\
\cellcolor{blueBg} & \cellcolor{blueBg}Structural Enclosure & \cellcolor{blueBg}Count objects hidden within enclosed or covered spaces & \cellcolor{blueBg}Objects in the microwave? \\
\midrule

\cellcolor{greenBg}\textcolor{esiGreen}{\textbf{Temporal}} & \cellcolor{greenBg}Unobserved Change & \cellcolor{greenBg}Infer scene changes that occurred during an unobserved interval & \cellcolor{greenBg}What changed when I looked away? \\
\cellcolor{greenBg}\textcolor{esiGreen}{\textbf{Understanding}} & \cellcolor{greenBg}Agent Observation & \cellcolor{greenBg}Reason about scene dynamics from other agents & \cellcolor{greenBg}Other robot's world model? \\
\midrule

\cellcolor{greenBg}\textcolor{esiGreen}{\textbf{Action Sequencing}} & \cellcolor{greenBg}Action Order Inference & \cellcolor{greenBg}Determine the correct procedural ordering of a sequence of actions & \cellcolor{greenBg}What is the action sequence for assembly? \\

\bottomrule
\end{tabular}
\end{table*}

\section{Per-Category Task Distribution}
\begin{figure*}[htbp]
    \centering
    \includegraphics[width=\textwidth]{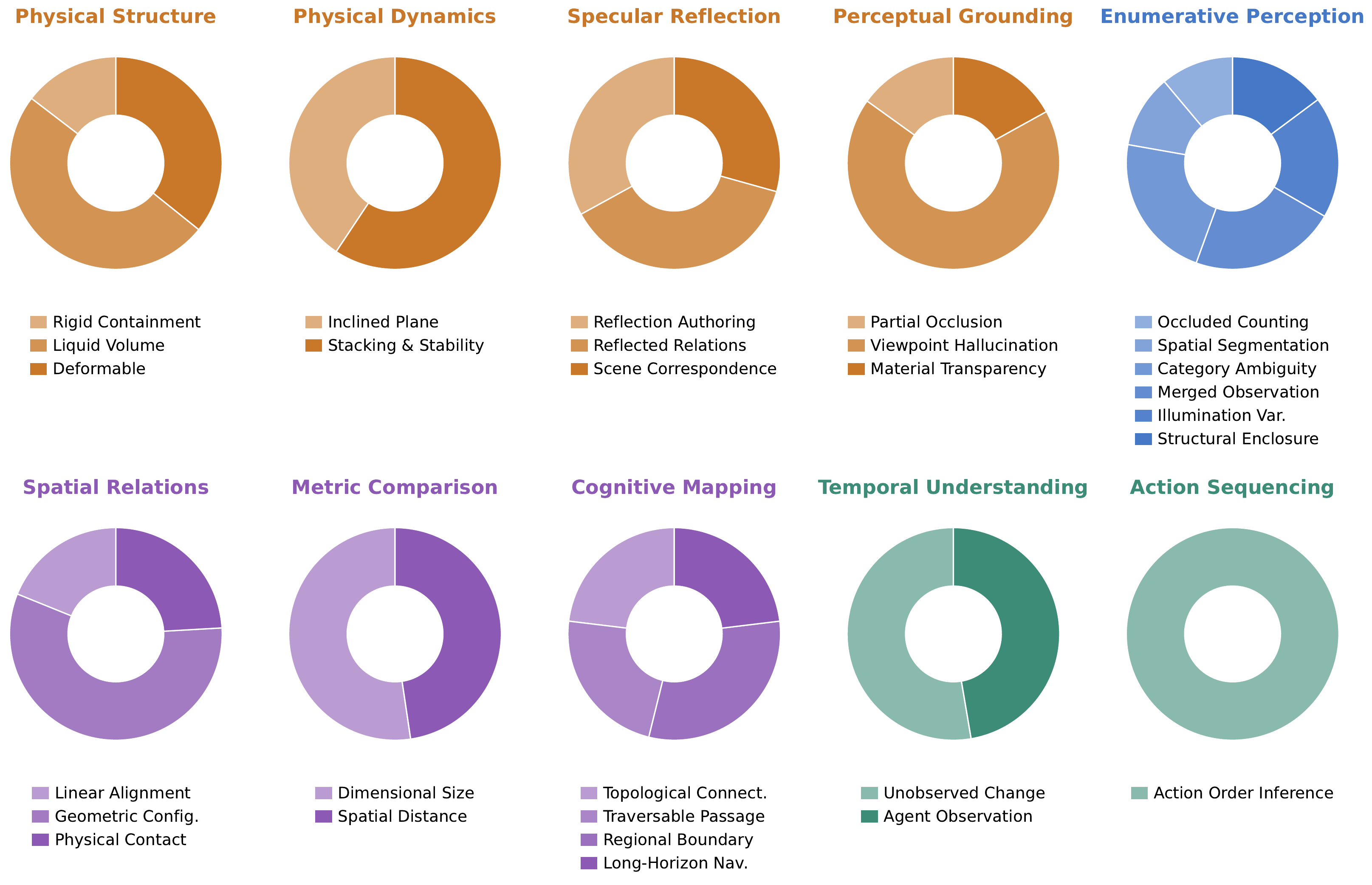}
    \caption{Subcategory distribution within each of the 10 \textsc{ESI-Bench} task categories.}
    \label{fig:subcategory_distribution}
\end{figure*}

In Figure~\ref{fig:subcategory_distribution}, we show the detailed distribution of sub-categories inside each cateogyr.

\section{Human Study Protocol}
\label{appendix:human_study}

\paragraph{Participants and compensation.}
We recruited 28 participants via internal volunteer sign-up, all with normal
or corrected-to-normal vision, and all was not exposed to such tasks before, ensuring fairness. Each participant was compensated at a rate
above local minimum wage. The study was conducted in accordance with
institutional guidelines; no IRB approval was required as the study involved
no sensitive data collection and participants interacted only with simulated
environments. The participants evaluate on the full benchmark like our models. 

\paragraph{Interface.}
Participants interacted with a web-based interface displaying the egocentric
view of an agent inside an OmniGibson scene. At each step, the participant
selected one action from the same discrete action vocabulary available to
evaluated models: locomotion actions (\texttt{move\_forward},
\texttt{move\_backward}, \texttt{move\_left}, \texttt{move\_right},
\texttt{move\_up}, \texttt{move\_down}), perception actions
(\texttt{turn\_left}, \texttt{turn\_right}, \texttt{turn\_up},
\texttt{turn\_down}), manipulation actions (\texttt{pick\_up},
\texttt{put\_inside}, \texttt{put\_on}, \texttt{fill\_with\_water},
\texttt{pour}), and the terminal action $\texttt{answer}(\hat{y}, c)$
which commits to a final free-form answer with an associated confidence
score $c \in [0, 1]$.
The step budget was capped at $T_{\max} = 30$, matching the budget
available to all evaluated models. We strictly guarantee that the human interface is exactly the same as active model interface with the same pixel-level egocentric rendering and action discretization affects. 

\paragraph{Task assignment.}
Each participant completed a subset of tasks drawn from at least 6 of the
10 \textsc{ESI-Bench} categories, with tasks sampled to balance category coverage
across participants. Each task instance was completed by at least 5
participants independently; disagreements on the ground-truth answer were
resolved by a third annotator.

\paragraph{Instructions.}
Participants were shown a brief written tutorial with two worked examples
before beginning. They were instructed to explore freely, gather as much
evidence as they deemed necessary, and answer only when confident. No hints
about the correct answer or the nature of the spatial challenge were
provided.

\paragraph{Data collected.}
For each task instance and participant we recorded: the full action sequence
and egocentric frame at each step, the final answer and confidence score,
the step at which \texttt{answer} was issued, and any intermediate
confidence ratings logged during exploration.

\section{More Benchmark Examples}
\begin{figure}[htbp]
    \centering
    \includegraphics[width=\textwidth]{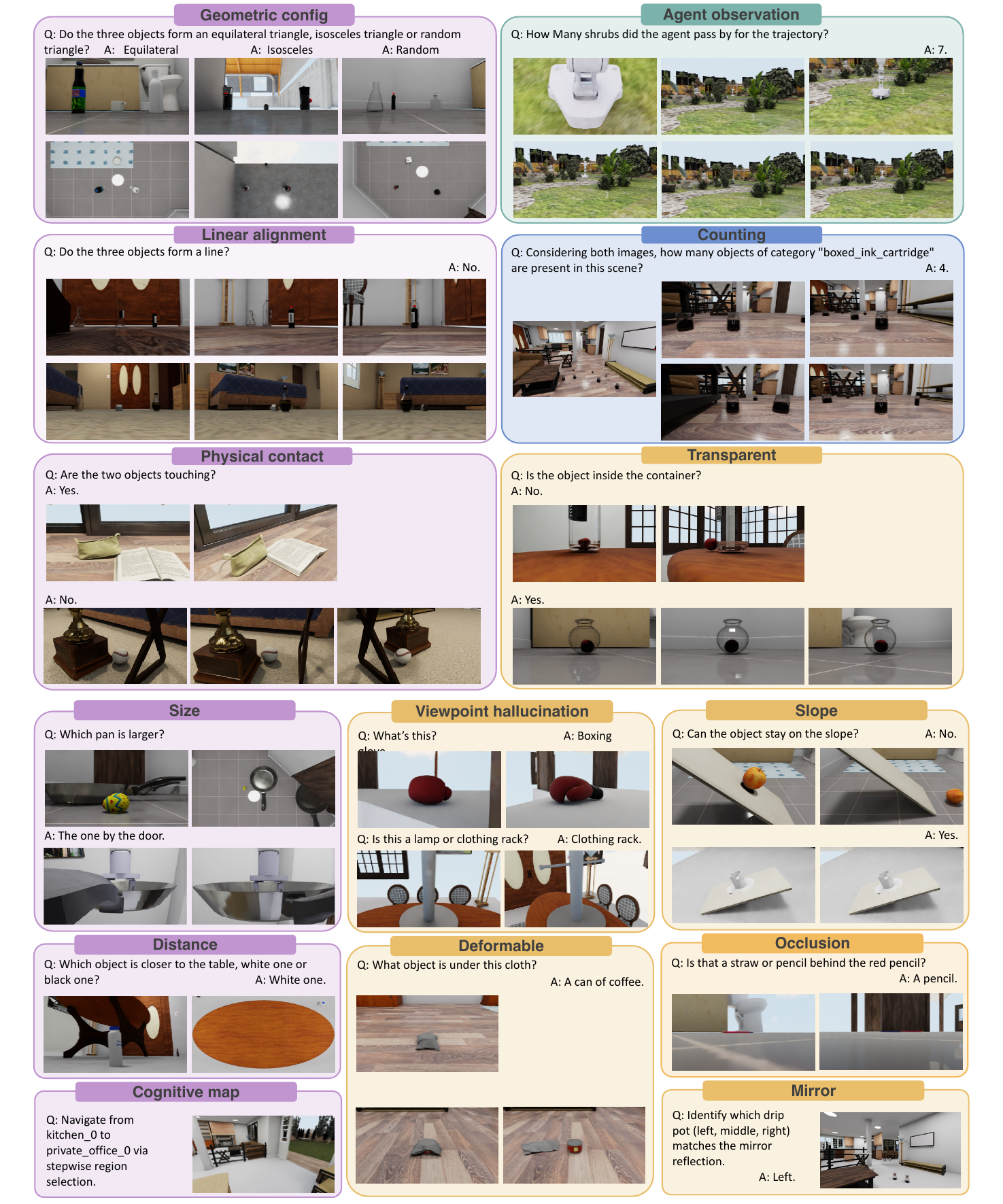}
    \caption{\small Additional benchmark examples from \textsc{ESI-Bench}, organized by core knowledge systems: \textcolor{esiOrange}{object representation}, \textcolor{esiPurple}{layout and geometry}, \textcolor{esiBlue}{number representation}, and \textcolor{esiGreen}{agents and goal-directed actions}. }
    \label{fig:benchmark_examples}
\end{figure}

In figure~\ref{fig:benchmark_examples}, we show more data points / examples of our \textsc{ESI-Bench} datasets.

\section{More Qualitative Examples}
\label{more_qualitative}
\begin{figure}[H]
    \centering
    \includegraphics[width=\textwidth]{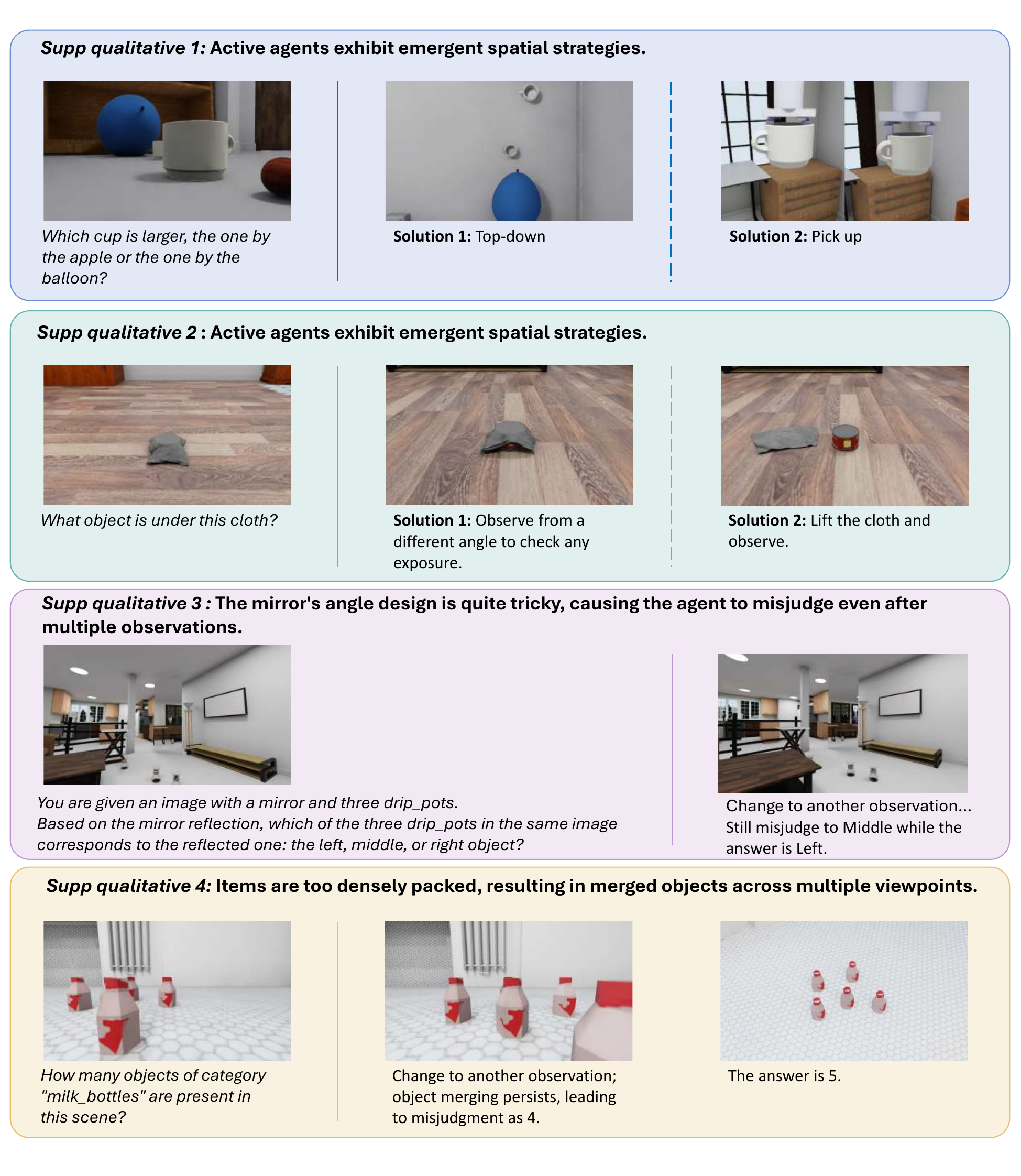}
    \caption{Additional qualitative examples illustrating emergent agent behaviors and failure modes: (1)~active agents employ emergent spatial strategies such as top-down viewing and object manipulation to compare sizes; (2)~active agents explore alternative viewpoints or interact with the environment to identify hidden objects; (3)~tricky mirror angle designs cause persistent misjudgment; (4)~densely packed objects lead to merged perception across viewpoints, resulting in undercounting.}
    \label{fig:qualitative_examples}
\end{figure}

\section{Human Verification and Generator Bias Analysis}
\label{app:human_verification_generator_bias}

This appendix provides additional details on human verification and on our analysis of possible generator bias. GPT-4o is used only as a scalable proposal engine during dataset construction: it proposes candidate objects, object placements, initial agent poses, natural-language questions, and candidate action trajectories from simulator scene graphs. The final benchmark instances, however, are not accepted directly from GPT-4o. Each candidate is instantiated and executed in OmniGibson, filtered through simulator-state checks, and then verified by human annotators using rendered observations and simulator metadata. This ensures that the final benchmark is grounded in physical simulation and human judgment rather than relying on GPT-4o as an automatic source of correctness.

Because LLM-assisted benchmark construction may introduce systematic artifacts, we further audit GPT-4o-generated tasks for possible linguistic and object-category. The purpose of this analysis is to test whether models could solve \textsc{ESI-Bench} by exploiting superficial regularities rather than actively acquiring task-relevant evidence. We evaluate shortcut baselines that remove visual observations and action histories, measure answer/object/template diversity, and compare model accuracy on matched GPT-4o-generated and human-generated task subsets.

\subsection{Human Verification Protocol}
\label{app:human_verification_protocol}

After simulator-based filtering, every candidate task is reviewed by human annotators. Human verification is designed to ensure that each retained instance is \textbf{correct}, \textbf{answerable}, and \textbf{non-trivial}. Annotators are not asked to judge whether a task is interesting in the abstract; instead, they follow a structured checklist tied to the simulator state, rendered observations, and the available action space.

For each candidate task, annotators are shown: (1) the natural-language question and proposed ground-truth answer; (2) the initial egocentric observation; (3) rendered observations along the verified trajectory; (4) the final simulator state after executing the trajectory; and (5) task metadata, including object categories, object states, spatial relations, containment/contact flags, visibility masks, and room assignments when needed. This allows annotators to verify both the visual evidence available to an agent and the simulator-derived state used to determine the answer.

Annotators evaluate each instance along three axes.

\paragraph{Correctness.}
Correctness measures whether the task is physically and semantically consistent with the simulator state. Annotators verify that the relevant objects are successfully instantiated, that object placements are physically plausible, that the proposed trajectory can be executed without invalid transitions, and that the proposed answer matches the simulator-derived state. For example, a containment task is marked correct only if the target object is actually inside the container according to simulator state and the rendered observations are consistent with this state. A contact task is marked correct only if the contact relation is supported by simulator contact flags and is visually plausible in the rendered scene. Instances are rejected if objects float, intersect unnaturally, disappear, fail to settle, or if the proposed answer contradicts the simulator state.

\paragraph{Answerability.}
Answerability measures whether the question has a unique answer that can be obtained through the available action space. Annotators verify that the required evidence can be acquired through locomotion, viewpoint control, or manipulation within the step budget. A task is rejected if the relevant object is never observable, if the answer depends on subjective interpretation, if multiple answers are equally plausible, or if the available actions cannot reveal the necessary evidence. For example, a question about whether an object is inside a closed container is answerable only if the agent can inspect, open, move, pour, or otherwise interact with the relevant objects in a way that reveals the hidden state.

\paragraph{Non-triviality.}
Non-triviality measures whether the task genuinely requires embodied evidence acquisition. Annotators reject tasks that can be answered from the initial observation alone, from obvious category priors, or from superficial visual cues unrelated to the intended spatial reasoning. For example, an occlusion task is rejected if the hidden object is already fully visible at a perfect angle in the first frame. The goal is to retain tasks where the agent must actively decide what information to acquire.

Each instance is independently reviewed by 3 annotators. For each verification axis, annotators assign a binary pass/fail label and may provide a short rejection reason. An instance is retained only if it passes all three axes. When annotators disagree, the disagreement is resolved by majority vote. Rejected instances are removed from the benchmark and are not used in evaluation.

\subsection{Verification Scores for GPT-4o-Generated Tasks}
\label{app:gpt4o_verification_scores}

Table~\ref{tab:gpt4o_human_verification} reports category-level human verification scores for GPT-4o-generated task proposals after simulator filtering. Correctness, answerability, and non-triviality are reported separately, and \textit{Overall Pass} denotes the percentage of instances that pass all three checks. The high verification scores indicate that GPT-4o-generated proposals are generally consistent with simulator state, answerable through embodied interaction, and non-trivial after the validation pipeline.

\begin{table}[h]
\centering
\small
\setlength{\tabcolsep}{5pt}
\begin{tabular}{lcccc}
\toprule
\textbf{Category} & \textbf{Correctness} & \textbf{Answerability} & \textbf{Non-triviality} & \textbf{Overall Pass} \\
\midrule
Perceptual Grounding      & 94.7 & 93.8 & 91.6 & 88.9 \\
Physical Structure        & 91.3 & 90.5 & 88.2 & 84.6 \\
Physical Dynamics         & 89.8 & 88.7 & 86.9 & 82.1 \\
Specular Reflection       & 93.4 & 92.1 & 89.5 & 86.7 \\
Spatial Relations         & 92.6 & 91.4 & 87.8 & 85.2 \\
Metric Comparison         & 95.1 & 94.3 & 90.7 & 89.4 \\
Cognitive Mapping         & 96.4 & 95.8 & 92.6 & 91.2 \\
Enumerative Perception    & 87.9 & 86.5 & 83.9 & 78.6 \\
Temporal Understanding    & 90.7 & 89.9 & 87.1 & 83.6 \\
Action Sequencing         & 88.5 & 87.3 & 85.4 & 80.9 \\
\midrule
Average                   & 92.0 & 91.0 & 88.4 & 85.1 \\
\bottomrule
\end{tabular}
\caption{Human verification scores for GPT-4o-generated task proposals after simulator filtering. Correctness measures physical and semantic consistency with the simulator state; answerability measures whether the question has a unique answer obtainable through the available action space; non-triviality measures whether the task requires embodied evidence acquisition rather than being solvable from the initial observation or priors. Overall pass denotes the percentage of instances that pass all three checks.}
\label{tab:gpt4o_human_verification}
\end{table}

The verification scores also reveal where automatic task proposal is most challenging. Categories involving physical interaction and hidden state changes, such as Physical Dynamics, Enumerative Perception, Temporal Understanding, and Action Sequencing, have lower overall pass rates than categories such as Cognitive Mapping and Metric Comparison. This is expected because these categories require more precise control over physical state, occlusion, object interaction, or temporal dependencies. Human verification therefore plays an important role in removing invalid, ambiguous, or insufficiently interactive instances before evaluation.

\subsection{Bias Audit for GPT-4o-Generated Tasks}
\label{app:gpt4o_bias_audit}

A central concern for LLM-assisted benchmark construction is that generated tasks may contain systematic artifacts. In \textsc{ESI-Bench}, such artifacts could appear in two forms. First, \textbf{linguistic bias}: question wording might reveal the answer. Second, \textbf{object-category bias}: particular object categories might correlate with particular labels. We therefore evaluate shortcut baselines and dataset-level diversity statistics.

\paragraph{Linguistic bias.}
We test whether task questions alone contain enough information to predict the answer. For each task, we remove all visual observations, simulator metadata, and action history, and provide only the natural-language question.  If GPT-4o-generated questions contained strong linguistic artifacts, question-only accuracy would be substantially above majority baselines. Instead, question-only performance remains close to chance and below passive perception, indicating that question wording alone is insufficient to solve the tasks.

\paragraph{Object-category bias.}
We next test whether task-relevant object categories reveal the answer. We provide the model with the question and the unordered list of task-relevant object categories, but no object positions, rendered observations, action history, or simulator state. This metadata-only setting tests whether certain objects are systematically associated with particular answers. Metadata-only performance is slightly higher than question-only performance, as expected, because object categories provide weak priors about plausible physical interactions. However, it remains below passive single-view performance and far below active exploration, indicating that object identity alone does not provide a reliable shortcut.

\begin{table}[h]
\centering
\small
\setlength{\tabcolsep}{5pt}
\begin{tabular}{lcccc}
\toprule
\textbf{Category} 
& \textbf{Question-Only} 
& \textbf{Metadata-Only} 
& \textbf{Passive Single} 
& \textbf{Active} \\
\midrule
Perceptual Grounding      & 26.1 & 28.3 & 32.8 & 66.3 \\
Physical Structure        & 31.5 & 35.0 & 50.7 & 64.1 \\
Physical Dynamics         & 49.2 & 51.0 & 51.9 & 81.2 \\
Specular Reflection       & 47.2 & 49.5 & 47.2 & 50.4 \\
Spatial Relations         & 30.8 & 35.1 & 45.4 & 54.8 \\
Metric Comparison         & 49.2 & 50.8 & 48.5 & 64.1 \\
Cognitive Mapping         & 47.6 & 50.1 & 53.9 & 57.9 \\
Enumerative Perception    & 24.3 & 26.1 & 13.0 & 22.6 \\
Temporal Understanding    & 35.7 & 38.1 & 37.4 & 53.0 \\
Action Sequencing         & 24.8 & 29.3 & 44.2 & 54.5 \\
\midrule
Average                   & 36.6 & 39.3 & 42.5 & 56.9 \\
\bottomrule
\end{tabular}
\caption{Shortcut baseline analysis for GPT-4o-generated tasks. Question-only uses only the natural-language question. Metadata-only uses the question and unordered task-relevant object categories, but no observations, positions, action history, or simulator state. Passive Single and Active report Gemini 3.1 category-level averages from the main evaluation. Shortcut baselines remain below active exploration, suggesting that benchmark performance is not explained by linguistic or object-category artifacts.}
\label{tab:gpt4o_bias_shortcuts}
\end{table}
Table~\ref{tab:gpt4o_bias_shortcuts} shows a clear separation between shortcut baselines, passive perception, and active exploration. The average question-only accuracy is 36.6\%, and metadata-only accuracy is 39.3\%, both below passive single-view accuracy at 42.5\%. Active exploration further improves to 56.9\%. This ordering indicates that language and object-category priors do not explain the benchmark results: visual evidence is needed even for passive performance, and action-guided evidence acquisition provides an additional gain beyond passive observation. The only exception is Enumerative Perception, where passive single-view is especially low because counting tasks often require resolving occlusion, enclosure, or merged observations; nevertheless, active exploration still substantially outperforms all shortcut baselines.

\begin{table}[h]
\centering
\small
\setlength{\tabcolsep}{5pt}
\begin{tabular}{lcc}
\toprule
\textbf{Category} & \textbf{Answer Balance} & \textbf{Object Diversity} \\
\midrule
Perceptual Grounding      & 0.92 & 0.87 \\
Physical Structure        & 0.89 & 0.84 \\
Physical Dynamics         & 0.95 & 0.82 \\
Specular Reflection       & 0.93 & 0.85 \\
Spatial Relations         & 0.90 & 0.88 \\
Metric Comparison         & 0.96 & 0.86 \\
Cognitive Mapping         & 0.94 & 0.80 \\
Enumerative Perception    & 0.88 & 0.89 \\
Temporal Understanding    & 0.91 & 0.83 \\
Action Sequencing         & 0.87 & 0.81 \\
\midrule
Average                   & 0.92 & 0.85 \\
\bottomrule
\end{tabular}
\caption{Bias and diversity statistics for GPT-4o-generated tasks. Answer balance is normalized answer entropy; object diversity is normalized entropy over task-relevant object categories. Higher values indicate less concentration around a single answer or object set.}
\label{tab:gpt4o_bias_diversity}
\end{table}

Table~\ref{tab:gpt4o_bias_diversity} further shows that GPT-4o-generated tasks are not dominated by a small set of labels, objects, or spatial templates. Answer distributions are balanced across categories, with an average normalized entropy of 0.92. Object diversity is also high, reflecting the use of BEHAVIOR-1K object inventories, random 200-category subsampling, and category-specific filtering.  However, even these categories retain substantial variation in relative placement and action requirements.

\subsection{Comparison with Human-Generated Tasks}
\label{app:gpt4o_human_generated_comparison}

As an additional check, we compare GPT-4o-generated tasks with a matched subset of human-generated tasks. Human annotators are given the same category definitions, action space, and simulator scene context, and are asked to author natural-language questions that require embodied spatial reasoning. We then construct matched GPT-4o-generated and human-generated subsets with the same category distribution and evaluate representative agents on both.

The purpose of this comparison is not to use human-generated tasks as an additional benchmark split, but to test whether GPT-4o-generated tasks introduce a systematic difficulty bias. If GPT-4o-generated tasks were substantially easier or harder than human-authored ones, we would expect large performance differences between the two subsets. Instead, as shown in Table~\ref{tab:gpt4o_human_accuracy_comparison}, model accuracies are broadly similar across task sources, with average gaps within a few percentage points.

\begin{table}[h]
\centering
\small
\setlength{\tabcolsep}{5pt}
\begin{tabular}{lccc}
\toprule
\textbf{Model} & \textbf{GPT-4o-Generated} & \textbf{Human-Generated} & \textbf{Absolute Gap} \\
\midrule
GPT-5 Active               & 51.6 & 48.9 & 2.7 \\
Gemini 3.1 Active          & 56.9 & 53.4 & 3.5 \\
VGGT + Gemini Active       & 51.6 & 48.1 & 3.5 \\
GT 3D + Gemini Active      & 68.0 & 63.6 & 4.4 \\
Human Active               & 75.3 & 72.1 & 3.2 \\
\midrule
Average                    & 60.7 & 57.2 & 3.5 \\
\bottomrule
\end{tabular}
\caption{Accuracy comparison between GPT-4o-generated and human-generated task subsets under active exploration. The gaps remain within a few percentage points, suggesting that GPT-4o-generated tasks have similar difficulty to human-generated tasks.}
\label{tab:gpt4o_human_accuracy_comparison}
\end{table}

Overall, these results support three conclusions. First, GPT-4o-generated task proposals achieve high human verification scores after simulator filtering, indicating that they are generally correct, answerable, and non-trivial. Second, question-only and metadata-only baselines remain below passive single-view performance and far below active exploration, suggesting that benchmark performance cannot be explained by linguistic or object-category shortcuts. Third, model accuracies on GPT-4o-generated and human-generated subsets are similar, with differences mostly within a few percentage points. This suggests that GPT-4o can serve as a scalable proposal mechanism without substantially distorting benchmark difficulty, provided that generated tasks are filtered through simulator execution and human verification.

\section{Diverse Emergent Behaviors in Active Exploration}
\label{sec:emergent_behaviors}

Active exploration not only improves accuracy, but also reveals that agents often develop diverse
ways of resolving the same type of spatial uncertainty. Since \textsc{ESI-Bench} does not provide
category-specific exploration policies, prescribed trajectories, or step-by-step instructions, these
behaviors are not hand-coded. They emerge from the interaction between the question, the current
egocentric observation, and the embodied action space.

To quantify this diversity, we conduct a human annotation study over successful active trajectories.
For each representative subcategory, annotators group trajectories according to their high-level
behavioral pattern, without being given predefined strategy labels. We then report the number of
distinct strategy clusters that emerge within each subcategory. A larger number indicates that agents
found multiple qualitatively different behavioral routes to acquire the evidence needed for the same
class of question.

\begin{table}[t]
\centering
\small
\setlength{\tabcolsep}{8pt}
\renewcommand{\arraystretch}{1.12}
\begin{tabular}{lc}
\toprule
\textbf{Subcategory} & \textbf{\# Emergent Strategy Clusters} \\
\midrule
Partial Occlusion         & 4 \\
Material Transparency     & 6 \\
Rigid Containment         & 3 \\
Liquid Volume             & 3 \\
Physical Contact          & 4 \\
Spatial Distance          & 3 \\
Dimensional Size             & 4 \\
\bottomrule
\end{tabular}
\caption{Human-annotated diversity of emergent behaviors in successful active trajectories.
For each representative subcategory, annotators group trajectories into distinct behavioral clusters.
The table reports the number of clusters discovered, without imposing predefined strategy labels.}
\label{tab:emergent_behavior_diversity}
\end{table}

Table~\ref{tab:emergent_behavior_diversity} shows that emergent behavior diversity varies
substantially across subcategories. They admit multiple possible routes to the answer: the agent can acquire
evidence through different viewpoints, different interactions, or different action orders. This suggests
that active exploration is not merely executing a single canonical policy, but can produce several
self-developed behavioral patterns for the same underlying spatial problem.

This analysis complements the aggregate accuracy results. Accuracy shows that active exploration
helps; strategy clustering shows that the improvement is not simply due to receiving more images.
Agents can develop multiple behavioral routes for acquiring task-relevant evidence, even without
being told how to explore. At the same time, diversity alone does not imply human-like epistemic
control. As discussed in Experiments section, models may still commit prematurely or fail
to revise beliefs under contradiction. 

\section{Failure Analysis of VGGT-Based 3D Augmentation}
\label{app:vggt_failure_analysis}

The VGGT-based 3D augmentation results raise an important question: why can reconstructed 3D scene graphs hurt performance even when explicit 3D information should in principle help? In our experiments, \texttt{VGGT + Gemini} underperforms the 2D Gemini baseline on several fine-grained spatial tasks, with the most striking case appearing in Counting w Occlusion. We attribute this degradation to errors introduced during 3D reconstruction and scene-graph construction, which can corrupt the symbolic input provided to the language model.

We manually inspect representative \texttt{VGGT + Gemini} failures and observe three recurring error modes. The dominant failure is \textbf{object duplication}: partially observed or depth-ambiguous objects are reconstructed as multiple fragments or repeated instances, causing the downstream scene graph to over-count objects. This is especially harmful for Enumerative Perception tasks, where the final answer depends directly on the number of object instances. A second failure mode is \textbf{object hallucination}: noisy geometry occasionally produces spurious object proposals or assigns object labels to background fragments. A third failure mode is \textbf{spatial-relation corruption}: inaccurate depth estimates distort relative positions, contact relations, containment, and near/far comparisons, leading the language model to reason over an incorrect spatial structure.

\begin{table}[h]
\centering
\small
\setlength{\tabcolsep}{5pt}
\renewcommand{\arraystretch}{1.08}
\begin{tabularx}{\linewidth}{p{2.7cm}X p{3.1cm}}
\toprule
\textbf{Failure Mode} & \textbf{Main Effect} & \textbf{Affected Tasks} \\
\midrule
Object duplication
& One partially observed object is reconstructed as multiple fragments or repeated instances, causing over-counting.
& Counting w Occlusion; Merged Observation \\

Object hallucination
& Noisy geometry or background fragments are converted into spurious object proposals.
& Enumerative Perception; Perceptual Grounding \\

Spatial-relation corruption
& Depth errors distort relative position, contact, containment, and near/far relations.
& Spatial Relations; Metric Comparison; Physical Structure \\
\bottomrule
\end{tabularx}
\caption{Common failure modes of VGGT-based 3D augmentation. The dominant issue is object duplication from noisy depth and partial reconstruction, followed by object hallucination and corrupted spatial relations.}
\label{tab:vggt_failure_modes}
\end{table}

These errors suggest that the poor performance of \texttt{VGGT + Gemini} is not caused by 3D information being unhelpful in general. Rather, it reflects a mismatch between the precision required by \textsc{ESI-Bench} and the reliability of current reconstructed scene graphs. Ground-truth 3D representations improve performance on several depth-sensitive tasks, showing that accurate 3D structure can be beneficial. However, when reconstruction is noisy, the resulting scene graph can become more misleading than a 2D observation, because the language model tends to treat the graph as a reliable symbolic description of the scene. This highlights a key challenge for 3D-augmented MLLMs: improving not only reconstruction quality, but also uncertainty-aware scene-graph construction that can distinguish reliable 3D evidence from ambiguous or fragmented geometry.
\
\section{Operationalizing the Metacognitive Gap}
\label{app:metacognitive_gap}

Our main experiments identify a metacognitive gap between humans and current MLLMs: humans tend to seek falsifying viewpoints, delay commitment under ambiguity, and revise beliefs when contradicted, whereas models often stop early with high confidence. In the main paper, we analyze this gap qualitatively through representative trajectories. Here we clarify how this behavior can be operationalized using trajectory-level measures.

We define three measurable axes of metacognitive behavior. First, \textbf{evidence sufficiency} measures whether the agent gathers sufficiently diverse observations before answering. This can be approximated by the diversity of viewpoints visited and whether the agent acquires observations that directly reveal the task-relevant evidence. Second, \textbf{falsification seeking} measures whether the agent selects actions that could disconfirm its current hypothesis, rather than only actions that preserve or confirm the initial belief. This can be estimated by annotating whether a new view is redundant, confirmatory, or contrastive with respect to the previous observation. Third, \textbf{belief revision} measures whether the agent changes its answer after later observations contradict earlier impressions.

\begin{table}[h]
\centering
\small
\setlength{\tabcolsep}{5pt}
\renewcommand{\arraystretch}{1.08}
\begin{tabular}{p{2.6cm}p{5.6cm}p{2.2cm}}
\toprule
\textbf{Axis} & \textbf{Definition} & \textbf{Signal} \\
\midrule
Evidence sufficiency 
& Whether the agent gathers sufficiently diverse observations before answering. 
& View diversity \\

Falsification seeking 
& Whether actions seek evidence that could disconfirm the current hypothesis. 
& Contrastive views \\

Belief revision 
& Whether the agent changes its answer after contradictory evidence. 
& Answer updates \\
\bottomrule
\end{tabular}
\caption{Operational measures for the metacognitive gap observed in \textsc{ESI-Bench}.}
\label{tab:metacognitive_axes}
\end{table}

We further compute these measures on active trajectories. \textbf{View diversity} is the percentage of trajectories containing observations from multiple meaningfully distinct viewpoints. \textbf{Contrastive view rate} is the percentage of trajectories containing at least one observation that disambiguates between competing hypotheses or reveals evidence against the initial hypothesis. \textbf{Belief revision rate} is the percentage of trajectories where the predicted answer changes after additional observations.

\begin{table}[h]
\centering
\small
\setlength{\tabcolsep}{5pt}
\renewcommand{\arraystretch}{1.08}
\begin{tabular}{lccc}
\toprule
\textbf{Agent} & \textbf{View Diversity} & \textbf{Contrastive View Rate} & \textbf{Belief Revision Rate} \\
\midrule
Human Active        & 71.8 & 62.7 & 41.3 \\
Gemini 3.1 Active   & 43.5 & 31.5 & 18.9 \\
GPT-5 Active        & 39.2 & 28.7 & 16.4 \\
\bottomrule
\end{tabular}
\caption{Trajectory-level measures of the metacognitive gap. Humans acquire more diverse and contrastive evidence, and revise beliefs more often than current MLLMs. All values are percentages.}
\label{tab:metacognitive_quantitative}
\end{table}

Table~\ref{tab:metacognitive_quantitative} shows that humans acquire more diverse viewpoints, seek contrastive evidence more often, and revise their beliefs more frequently after new observations. In contrast, models exhibit lower rates of falsification-seeking behavior and belief revision. These measures are complementary to accuracy: accuracy captures whether the final answer is correct, while metacognitive metrics capture whether the agent knows when its evidence is insufficient and how to acquire better evidence. A full benchmark for metacognition is beyond the scope of this work, but \textsc{ESI-Bench} provides the interaction traces needed to study these behaviors systematically.

\section{Analysis of Passive Multi-View Evaluation}
\label{app:passive_multiview_analysis}

Our experiments show that passive multi-view input often fails to improve over passive single-view input, and in some categories even degrades performance. This finding should be interpreted carefully. The passive multi-view baseline uses 30 randomly sampled views to match the active exploration step budget, but it is not intended to represent an optimal multi-view reasoning system. Instead, it asks whether simply providing more unselected observations is sufficient to close the gap between passive perception and active evidence acquisition.

The degradation can arise from several sources. First, \textbf{view selection randomness}: random views may miss the diagnostic region while still adding many irrelevant images. Second, \textbf{image overload}: current MLLMs may struggle to integrate 30 views, especially when only a few contain task-relevant evidence. Third, \textbf{lack of temporal grounding}: passive views are unordered and not tied to the agent's own actions, making it harder to infer how observations relate to spatial movement. Fourth, \textbf{conflicting evidence}: redundant or partially occluded views may introduce apparent contradictions that distract the model from the useful observation.

\begin{table}[h]
\centering
\small
\setlength{\tabcolsep}{5pt}
\renewcommand{\arraystretch}{1.08}
\begin{tabular}{p{3.5cm}p{6.8cm}}
\toprule
\textbf{Source} & \textbf{Possible Effect} \\
\midrule
Random view selection 
& Views may cover the scene broadly but miss the diagnostic evidence needed for the question. \\

Image overload 
& Many irrelevant images can dilute the few useful observations and increase reasoning burden. \\

No action grounding 
& Views are not connected to self-selected actions, making spatial integration harder. \\

Conflicting evidence 
& Occluded or partial views can introduce misleading cues across images. \\
\bottomrule
\end{tabular}
\caption{Possible reasons why passive multi-view input may fail to improve over passive single-view input.}
\label{tab:passive_multiview_failure_modes}
\end{table}

This analysis suggests that the weakness of passive multi-view should not be read as evidence that multi-view reasoning is inherently unhelpful. A stronger baseline could use learned view selection, oracle-selected diagnostic views, view ranking, or explicit cross-view spatial alignment. Our results instead show that unstructured view accumulation is not a substitute for active exploration: the agent must not only receive more observations, but decide which observations are worth acquiring.

\section{Why a Synthetic Simulator Benchmark?}
\label{app:synthetic_simulator}

\textsc{ESI-Bench} is built in simulation rather than collected directly in the real world because the goal of the benchmark is controlled diagnosis of embodied spatial intelligence. Many of our tasks depend on hidden physical or spatial variables---containment, contact, fill level, occlusion, room connectivity, object count, and unobserved state changes---whose ground truth must be known exactly. In real-world data, these variables are difficult to label exhaustively and often require additional sensors, manual inspection, or assumptions about the scene. In contrast, simulation provides direct access to object poses, segmentation masks, contact flags, containment relations, room assignments, and physical states, allowing us to construct unambiguous labels and reject invalid instances.

Simulation also allows us to control the source of difficulty. \textsc{ESI-Bench} is designed to test whether an agent knows \emph{what evidence to acquire}, not whether it can handle uncontrolled sensor noise, imperfect calibration, or real-world deployment constraints. By using OmniGibson and BEHAVIOR-1K scenes, we can systematically vary occlusion, viewpoint ambiguity, object placement, physical interaction, and action requirements while keeping the task answer verifiable. This makes it possible to separate perception errors from action-selection errors through diagnostic settings such as passive single-view, passive multi-view, active exploration, and ground-truth passive trajectories.

We do not claim that simulator performance directly transfers to real-world robotics. Rather, \textsc{ESI-Bench} serves as a controlled testbed for identifying failure modes that would be difficult to isolate in the real world. A real-world version would introduce additional challenges such as sensor noise, imperfect actuation, reconstruction errors, and environment variability. These are important directions, but they would obscure the central question studied here: whether current MLLMs can close the perception-action loop by actively selecting informative observations. The synthetic simulator setting therefore provides a necessary first step toward rigorous, reproducible, and fine-grained evaluation of embodied spatial intelligence.

\section{Active Agent Prompting and Action Selection}
\label{app:active_agent_prompting}

In the active exploration setting, the evaluated MLLM controls the agent through the same discrete action space used by human participants. At each timestep, the model receives the task question, the current egocentric observation, the previous action-observation history, and the full action vocabulary. It is then prompted to select exactly one next action. The selected action is executed in OmniGibson, producing the next egocentric observation. This loop continues until the model issues the terminal \texttt{answer($\hat{y}$, $c$)} action or reaches the maximum budget of $T_{\max}=30$ steps.

\paragraph{Action-selection prompt.}
We use the following prompt template for active exploration. The same template is used across task categories and models, with only the task question, observation history, and available objects filled in from the current instance.

\begin{tcolorbox}[colback=gray!5,colframe=gray!40,title=Active Exploration Prompt]
You are an embodied agent in a 3D environment. Your goal is to answer the question by actively exploring the scene.

You may take one action at each step. Use actions only when they help you gather evidence. Do not answer until you have enough evidence.

\textbf{Question:} \{question\}

\textbf{Current observation:} You are given the current egocentric image.

\textbf{Action history:} \{previous actions and brief observation summaries\}

\textbf{Available actions:}
\begin{itemize}
    \item \texttt{move\_forward}, \texttt{move\_backward}, \texttt{move\_left}, \texttt{move\_right}, \texttt{move\_up}, \texttt{move\_down}
    \item \texttt{turn\_left}, \texttt{turn\_right}, \texttt{turn\_up}, \texttt{turn\_down}
    \item \texttt{pick\_up(obj)}
    \item \texttt{put(obj, inside, obj)}
    \item \texttt{put(obj, on, obj)}
    \item \texttt{fill(obj, with\_water)}
    \item \texttt{pour(obj, from, obj, to, obj)}
    \item \texttt{answer(answer, confidence)}
\end{itemize}

Choose exactly one next action from the available action space.

Return your response in the following format:

\texttt{Reason:} briefly explain what evidence you need next.

\texttt{Action:} one valid action.
\end{tcolorbox}

\paragraph{Action parsing and execution.}
Model outputs are parsed into the nearest valid action string. If the output contains an invalid action, malformed object reference, or multiple actions, we reprompt the model once with the same observation and a reminder to output exactly one valid action. If the second response remains invalid, the step is counted as invalid and the agent remains in place. For manipulation actions, object names are matched to visible or previously observed object identifiers when possible. Failed physical executions, such as attempting to pick up an unreachable object or placing an object into an incompatible container, are executed as simulator failures and included in the trajectory, since such failures reflect the agent's action-selection behavior.

\section{Why High-Level Actions?}
\label{app:high_level_actions}

\textsc{ESI-Bench} uses a high-level discrete action space rather than low-level motor control because the benchmark is designed to diagnose embodied spatial reasoning, not visuomotor control. Many tasks require agents to decide \emph{what evidence to acquire}: whether to move to another viewpoint, inspect behind an occluder, manipulate a container, compare distances, or test a physical relation. These decisions operate at the level of information-seeking behavior. Introducing low-level control would add additional sources of failure, such as grasp instability, collision recovery, locomotion drift, controller tuning, and actuation noise, making it difficult to determine whether an error comes from spatial reasoning or motor execution.

The high-level action space therefore acts as a controlled abstraction. It preserves the core perception-action loop---agents still choose where to move, what to inspect, what to manipulate, and when to answer---while reducing confounds from robot-specific control policies. This makes the evaluation more diagnostic: when an agent fails, the failure is more likely to reflect poor evidence acquisition, action selection, belief updating, or spatial reasoning, rather than low-level controller failure.

We acknowledge that low-level control is an important dimension of embodied intelligence. However, it is orthogonal to the main question studied in \textsc{ESI-Bench}. A low-level version of the benchmark could evaluate whether agents can translate spatial plans into continuous motor commands, handle imperfect actuation, and recover from execution errors. We view this as a complementary direction. \textsc{ESI-Bench} intentionally factors out low-level control so that the benchmark can focus on the higher-level spatial competence required to close the perception-action loop.

\section{Step Budget Ablation}
\label{app:step_budget_ablation}

We analyze how the maximum exploration budget affects active performance. Since active exploration gives the agent additional opportunities to acquire evidence, it is important to check whether performance continues to improve simply by allowing more steps. We therefore evaluate Gemini 3.1 Active with step budgets from 5 to 50, increasing the budget in intervals of 5 steps.

\begin{figure}[h]
\centering
\includegraphics[width=0.65\linewidth]{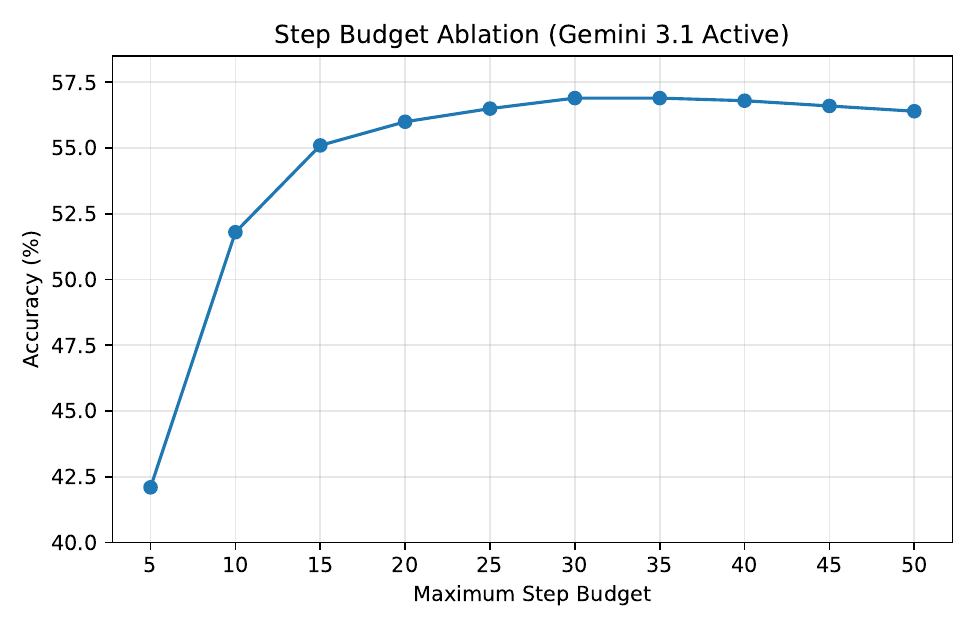}
\caption{Step budget ablation for Gemini 3.1 Active. Performance rises quickly up to 15--20 steps, saturates around 30 steps, and slightly decreases after 40 steps.}
\label{fig:step_budget_ablation}
\end{figure}

Figure~\ref{fig:step_budget_ablation} shows that performance improves rapidly from 5 to 15 steps, begins to saturate around 15--20 steps, and becomes nearly flat after 30 steps. Increasing the budget beyond 40 steps does not improve accuracy and even leads to a slight decrease. This suggests that the useful part of active exploration is not unbounded observation accumulation, but targeted evidence acquisition within the first few informative actions. Once the agent has acquired the relevant views or interactions, additional steps mostly introduce redundant or distracting observations.

This saturation pattern motivates our choice of $T_{\max}=30$ in the main experiments. A 30-step budget is large enough to support multi-step navigation, viewpoint adjustment, and manipulation, while avoiding unnecessarily long trajectories that add computational cost without improving performance. The slight degradation at larger budgets also suggests that longer exploration is not always beneficial: models may revisit redundant views, accumulate conflicting evidence, shift too much from task-relevant regions, or delay commitment even after sufficient information has already been observed.

\section{Per-Category Task Construction Details}
\label{sec:task_construction_details}

Each task category in \textsc{ESI-Bench} is instantiated through a dedicated pipeline
script that (i) selects task-relevant objects, (ii) places them in a
physically valid configuration, (iii) renders a fixed set of egocentric
camera views, and (iv) records ground-truth labels together with per-view
segmentation visibility flags.
All scripts share a common infrastructure: OmniGibson scenes are loaded
restricted to a single room, object categories are drawn from a 200-category
subsample of the BEHAVIOR-1K inventory, GPT-4o is queried for
task-appropriate category selection where semantic judgment is required,
and physics is settled for a fixed number of simulation steps before any
rendering or label extraction occurs. 
\paragraph{Initial Agent Viewpoint Setup.} The initial agent viewpoint for each category is constructed to be task-diagnostic: for containment tasks, the agent camera is placed collinear with the container and object so that the object is occluded behind or inside the container from view~0; for distance and size tasks, the initial azimuth is set along the direction of one of the objects or reference-anchored object, creating forced-perspective ambiguity; for contact tasks, the agent views the contact zone head-on where the gap is smallest; for stacking tasks, the initial agent sensor is placed at near-ground height where vertical ordering is hardest to judge; and for geometric configuration tasks, the initial view is taken from the centroid at grazing height where collinearity and triangle shape are indistinguishable. In each case the initial observation is deliberately insufficient to resolve the question, motivating the need for active exploration. \textbf{Instances where task-relevant objects are not visible in any rendered view under the per-view segmentation check described above are discarded. }

Below we describe the construction procedure for each of the ten task categories.

\subsection{Physical Structure: Material Transparency}
\label{sec:construction_transparent}

\paragraph{Object selection.}
A transparent container is sampled uniformly at random from a human-curated
list of container categories; its model instance is drawn from the BEHAVIOR-1K
inventory.
A small object is sampled from a separate human-curated list, with model
resolution via the inventory.

\paragraph{Scale.}
The small object is rescaled so that its largest bounding-box dimension equals
$\min(\text{Uniform}(l, u),\; 0.5 \times d_{\min}^{\text{container}})$
metres, where $d_{\min}^{\text{container}}$ is the smallest of the container's
three inventory bounding-box dimensions at unit scale, ensuring the object is
always physically smaller than the container's narrowest cross-section.

\paragraph{Placement.}
The container is placed on any surface object whose name contains ``table'',
falling back to the room floor.
For positive instances, containment is attempted via OmniGibson's
\texttt{Inside} object state; if that fails, the small object is teleported
to 25\% of the container's interior height and settled for 30 simulation
steps, with a bounding-box containment check applied.
If the directly behind position is occupied, up to 8 lateral offsets of
$0.05$ m each are tried with vacancy checks against all scene-graph bounding
boxes; a live-AABB correction pass then adjusts the final position so the
gap is exactly $0.01$ m.
Instances that fail placement at both attempts (positive) or all lateral
candidates (negative) are discarded.

\paragraph{Agent intitial spotion and view.}
For positive instances the initial azimuth is fixed along $-Y$; for negative
instances it is set opposite to the container-to-small-object axis, so that
view~0 always maximises occlusion of the small object behind the container.

\paragraph{Ground-truth label.}
Binary: \emph{inside} (positive) or \emph{not inside} (negative).
\subsection{Physical Structure: Liquid Volume}
\label{sec:construction_liquid}

\paragraph{Object selection.}
Fillable container pairs are selected from a human-precomputed manifest
such that their measured particle
capacities differ by at least 200 particles (\texttt{TARGET\_CAP\_DIFF}).
Capacities are measured offline via a water-dip pipeline: each container is
submerged in a particle-based water system simulated with GPU dynamics in BEHAVIOR, particle positions are clustered using
a connected-component graph over a KD-tree with radius equal to twice the
particle radius, and the largest connected component count is recorded as
the fillable capacity. 

\paragraph{Pouring Simulation.}
When water from one container is poured to another, we first fill that container with water particles using BEHAVIOR's dipping system. When it is poured into another container, we transfer the water particles from current container to target container, producing one of two
scenarios: complete transfer or
overflow.

\paragraph{Ground-truth label.}
The container with the larger offline-measured particle capacity.

\subsection{Physical Structure: Deformable}
\label{sec:physical_deformable}

\paragraph{Object selection.}
We first use GPT-4o to generate an object pool, and then filter it to retain portable objects with an average radius between \(5\) and \(15\,\mathrm{cm}\). For each question, we randomly sample one target object from this precomputed pool. Three distractor categories are sampled from the remaining categories to form a four-way multiple-choice question. We also sample one usable cloth asset with a side length no larger than \(50\,\mathrm{cm}\), prioritizing cloth models marked as recommended.

\paragraph{Object placement.}
The target object is placed on the selected room floor at a free location with a fixed clearance from nearby obstacles. Candidate positions are sampled from the traversability map when available, and otherwise from free floor locations. The object is dropped slightly above the floor and allowed to settle. The cloth is then positioned above the settled object center and dropped with a downward velocity until it covers the object.

\paragraph{Agent initial view.}
The main camera is placed at a fixed azimuth around the covered object and points toward the cloth-covered target. Its distance and height are adjusted according to the cloth footprint. We use a fixed field of view of \(70^\circ\).

\paragraph{Ground-truth label.}
The ground-truth label is the semantic category of the object hidden under the cloth. The answer choices contain the target category and three distractor categories, shuffled into options A--D.

\subsection{Physical Dynamics: Inclined Plane}
\label{sec:construction_slope}

\paragraph{Object selection.}
A room from the BEHAVIOR-1K scene pool is loaded; candidate object categories
are drawn from a room-object compatibility list, excluding floors,
ceilings, and walls.
A single object is selected and placed on a procedurally generated slope. \textbf{The mass of the object is randomly shifted from its original mass.}

\paragraph{Slope construction.}
A rigid box primitive with half-extents $(x, 0.15, 0.01)$ m is tilted at an
angle drawn uniformly from $[10^\circ, 45^\circ]$ from horizontal, where the slope base
half-extent $x$ is drawn uniformly from $[0.20, 0.30]$ m.
Static friction is drawn uniformly from $[0.3, 3.0]$; dynamic friction is
drawn uniformly from $[0.3, f_s]$ where $f_s$ is the sampled static friction,
ensuring the physical constraint $f_d \leq f_s$ is always satisfied.
Both coefficients are applied to the slope and object via an Isaac Sim physics
material with restitution $0.1$.
The physics timestep is $1/240$ s; action timestep is $1/60$ s.

\paragraph{Simulation and label derivation.} When the object is placed at the slope surface and the simulation runs for 30
steps. Slide is detected if the object's XY displacement exceeds 0.03 m;
fall is detected if the Z coordinate drops more than 0.05 m below the slope
bottom. The ground-truth label (stable / slides / falls) is derived from
these thresholds.

\paragraph{Agent initial viewpoint.}
The agent stands The initial view captures the object stands near the slope on the floor.

\paragraph{Ground-truth label.}
Stable or not.
\subsection{Physical Dynamics: Stacking and Stability}
\label{sec:construction_stacking}

\paragraph{Object selection.}
GPT-4o receives a 200-category randomly-sampled of the BEHAVIOR-1K inventory (different for every data point) and
selects 2 or 3 objects that are ``solid, not soft/flexible/liquid, and
plausibly stackable.''

\paragraph{Scale normalization.}
All selected objects are scaled to a uniform XY footprint equal to the
largest natural XY extent among them, with Z scaled proportionally.
This controls for trivial size mismatches and isolates shape-based stability.

\paragraph{Placement.}
For each permutation, the bottom object is placed on the floor via
\texttt{sample\_kinematics("onTop", floor)}, subsequent objects are placed
on top via the same primitive, and physics is settled for 60 steps.
Stability is then assessed along two axes.
First, an \texttt{is\_on\_top} check verifies that the upper object's AABB
minimum Z exceeds the lower object's AABB minimum Z by at least $0.02$ m,
and that the XY intersection-over-union of the two AABBs is at least $0.1$,
confirming the upper object is both elevated above and spatially overlapping
the lower one.
Second, a tilt check computes the dot product of the object's world-up vector
with the global $+Z$ axis; a stack layer is considered upright only if this
dot product exceeds $0.9$, corresponding to a tilt of less than roughly
$26^\circ$ from vertical.

\paragraph{Ground-truth label.}
The stable stacking order(s) among all permutations tried, derived from
\texttt{is\_on\_top()} outcomes after physics settling.

\subsection{Metric Comparison: Spatial Distance}
\label{sec:construction_distance}

\paragraph{Object selection.}
A reference object already present in the loaded room scene graph is selected
based on clearance: all four corners of its AABB are evaluated for clearance
to the nearest other object's bounding-box centre; at least two corners with
clearance $> 0.5$ m are required.
GPT-4o selects two similar-category objects to serve as the near and far
distractors.

\paragraph{Placement.}
The near object (\texttt{obj\_near}) is placed m
from one qualifying corner outward along the diagonal unit vector.
The far object (\texttt{obj\_far}) is placed with a slightly longer distance m
from a second qualifying corner.
Both objects use identity quaternion orientation.

\paragraph{Agent initial position and view.}
The agent is positioned that from its initial view, one object appears much larger than the other and blocking some view, making it hard to deduct distance.

\paragraph{Ground-truth label.}
\texttt{obj\_near} (which one is closer) or \texttt{obj\_far} (which one is far).

\subsection{Metric Comparison: Dimensional Size}
\label{sec:construction_size}

\paragraph{Object selection.}
GPT-4o selects a medium-sized household object category (not tiny, not large
furniture) from a 200-category random-sampled subsample (different each time).
Two instances of the same model are loaded: \texttt{task\_obj1} at scale 1.0
and \texttt{task\_obj2} at scale drawn uniformly from $[1.2, 1.3]$.

\paragraph{Reference objects.}
Two visually distinct reference objects are placed within $0.05$ m of each
task object's bounding-box boundary, with each reference at least $0.30$ m
from the other task object's boundary.
References serve as unambiguous spatial anchors differentiating the two
task objects.

\paragraph{Agent initial position and view.}
The agent is positioned that from its initial view, one object is much closer than the other and blocking some view, making it hard to deduct size.

\paragraph{Ground-truth label.}
The reference object to which the object is larger / samller.

\subsection{Spatial Relations: Physical Contact}
\label{sec:construction_touching}

\paragraph{Object selection.}
GPT-4o selects 2 medium-to-small-sized objects (e.g., teddy bear, book,
basketball) from a randomly-sampled 200-category subsample, avoiding tiny items (cups, pens)
and large furniture.

\paragraph{Placement.}
For positive (touching) instances, the two objects are placed in physical
contact: $\texttt{obj2\_xmin} = \texttt{obj1\_xmax}$ (zero X gap, faces
meeting) and $\texttt{obj2\_ymin} = \texttt{obj1\_ymax} - 0.05$ m
(confirming contact along Y). A small random X drift of $\pm 0.03$ m is
applied per instance for positional variety. After placement, a pixel-level
mask adjacency check via dilation overlap of the two instance segmentation
masks verifies that contact is visually confirmed.
For negative (non-touching) instances, the same initial placement is attempted
but an iterative correction loop pushes \texttt{obj2} outward in X until both
the X and Y face-to-face gaps exceed $0.05$ m, ensuring a clearly visible
separation. In both cases a live-AABB verification pass confirms the final gap
values before the instance is accepted.

\paragraph{Initial agent position and view.}
The agent is placed in a view which can be delusive (e.g., appears contacting when it's not).

\paragraph{Ground-truth label.}
Binary: the two objects are in contact or not.

\subsection{Spatial Relations: Linear Alignment and Geometric Configuration}
\label{sec:construction_line_triangle}

\paragraph{Object selection.}
For both Linear Alignment and Geometric Configuration, GPT-4o selects 3
small-footprint objects (table-top or hand-held size, with some Z height)
from a random-sampled 200-category subsample, avoiding large furniture, vehicles, and mats.

\paragraph{Linear Alignment placement.}
The three objects are placed collinearly along a random room axis, spaced
uniformly. Positive instances use strict collinearity; negative instances
introduce controlled lateral offsets that break alignment from most viewpoints.

\paragraph{Geometric Configuration placement.}
Objects are placed at the vertices of an equilateral triangle with side length
drawn uniformly from $[0.4, 0.65]$ m.
A size-compatibility check is applied: if the ratio of the largest to smallest
object XY footprint exceeds $2.0$, the object set is resampled (up to
5 attempts), preventing degenerate configurations where one object dominates
the triangle vertex.

\paragraph{Agent initial position and view.}
The agent is placed at ground level,
0° orbit from the centroid of the three objects, where collinearity and triangle shape are indistinguishable from a single frontal observation.

\paragraph{Ground-truth labels.}
Linear Alignment: binary (collinear / not collinear).
Geometric Configuration: categorical (equilateral / isosceles / random triangle).

\subsection{Physical Structure: Rigid Containment}
\label{sec:construction_storage}

\paragraph{Object selection.}
GPT-4o selects one fillable object category and three container categories of varying sizes (small, fit, large) from a 200-category subsample. Two additional non-fillable objects are independently sampled and scaled as a pair to jointly fill the large container, with each object's XY extent at the large container. With 50\% probability a third additional object is sampled and scaled to fit the small container, giving 3 or 4 objects to be assigned in total, each with a unique target container.

\paragraph{Pick-place simulation.}
Each object is placed into its target container via a two-stage kinematic pipeline. First, placement is attempted, settling physics for 30 steps and verifying success via a bounding-box containment check (object AABB fully within container AABB). If that fails, a fallback attempt uses  with the same settling and verification. A robot grasp-and-release primitive  is executed before each placement to simulate realistic pick-and-place manipulation. Robot eye-camera frames are captured before pickup, before placement, and after placement, providing egocentric views of each manipulation stage. 

\paragraph{Ground-truth label.}
The correct assignment of each object to its target container.

\subsection{Action Sequencing: Action Order Inference}
\label{sec:construction_action}

\paragraph{Object selection.}
GPT-4o designs a 2- or 3-object placement hierarchy
(e.g., plate $\rightarrow$ bowl $\rightarrow$ food item).
For each role in the hierarchy, several candidate instances are loaded: one that
clearly fits its target and one that barely fails, with scales chosen to make
the size difference physically meaningful but visually subtle. Others are confusables.

\paragraph{Layout.}
All objects are placed on the floor or table, with the fixed object at the centre and candidate pairs spread out left and right.

\paragraph{Ground-truth label.}
A spatial relation string specifying which candidate should be placed at each step of the correct action sequence.

\subsection{Temporal Understanding: Agent Observation}
\label{sec:construction_multiagent}

\paragraph{Object selection.}
GPT-4o selects 1--3 object categories to be placed as hidden objects along a robot navigation path within the scene.
\paragraph{Agent setup.}
Two agents operate simultaneously: a navigating robot that traverses a planned path through the scene, and a static observer agent fixed at the start position at 5.05.0
5.0 m height with ceiling removed, always looking at the navigating robot. The navigating robot's start and goal positions are sampled from traversable pixels of the BEHAVIOR-1K traversability map; the shortest path is validated to have at least 10 waypoints, with start/goal resampled up to 20 times if the dry-run reachability check fails.
\paragraph{Object placement.}
Hidden objects are placed at surface-appropriate positions along the confirmed path via \texttt{OnTop} state sampling, such that the navigating robot passes within 1.01.0
1.0 m of each.
\paragraph{Observer agent position and view.}
The static observer agent is fixed at the start position at 5.05.0
5.0 m height with ceiling removed, always oriented to face the navigating robot. It records what objects the navigating robot passes during its trajectory, serving as the ground-truth witness whose observations form the basis of the question posed to the evaluated model.
\paragraph{Ground-truth label.}
The count and categories of hidden objects the navigating robot passes within the proximity threshold, as observed by the static agent.

\subsection{Temporal Understanding: Unobserved Changes}
\label{sec:temporal_changes}

\paragraph{Object selection.}
We have three question types for this task: change detection, change identification and current-state reasoning. For each question, we sample one or more boxes depending on the task type. Change detection and change identification use a single box, while current-state reasoning uses three colored boxes. Box contents are sampled from a precomputed count-target object pool generated by GPT-4o. Depending on the sampled change type, the content may be kept the same, replaced by another category, removed, or added to an initially empty box.

\paragraph{Object placement.}
Boxes are placed on the selected room floor using the hidden-in-box placement routine. The placement respects the room bounding box, floor geometry, nearby blockers, and traversability constraints. For each phase, the corresponding content object is spawned and placed inside its assigned box; empty boxes contain no object. We involve two rendering phases to represent the time change. Phase 1 is rendered first, then its contents are removed and Phase 2 contents are spawned and rendered.

\paragraph{Agent initial view.}
Each phase is rendered from a primary viewpoint chosen to maximize visible boxes. Candidate camera positions are sampled around the room boundary and filtered by clearance and traversability. The selected camera looks toward the average box location.

\paragraph{Ground-truth label.}
The ground-truth label is derived directly from the sampled phase states. For change detection, the answer is whether any box changed. For change identification, the answer is the target box's change type: \texttt{replace}, \texttt{remove}, \texttt{add}, or \texttt{no\_change}. For current-state reasoning, the answer is the colored box whose content changed. Ground-truth close-up images are rendered by opening the boxes in each phase.

\subsection{Perceptual Grounding: Partial Occlusion}
\label{sec:construction_occlusion}

\paragraph{Object selection.}
A target object is sampled from a gpt-generated and human-curated manifest of occlusion-sensitive
categories and some confusable categories corresponding to the target categories when occluded. A compatible occluder is
sampled from a precomputed compatibility map,
ensuring the occluder is semantically plausible alongside the target.
The occluder is scaled per-axis so that each of its bounding-box dimensions
equals the target's corresponding dimension multiplied by a scale factor of
$1.1$, making it slightly larger than the target and thus capable of producing
meaningful partial occlusion.

\paragraph{Placement.}
Both objects are placed on a valid table surface via \texttt{OnTop} state
sampling. The occluder is positioned at a corner of the target's bounding box
so that from the canonical hard view it partially obscures the target.

\paragraph{Agent initial viewpoint.}
The agent is placed at $15^\circ$ lateral offset from the perfectly collinear
occluder-to-target axis, at the same height as the target centre, looking
horizontally at the target. This near-collinear position maximises ambiguity:
the target is partially hidden but not fully obscured, making it impossible
to identify with certainty from the initial view.

\paragraph{Ground-truth label.}
The category of the partially occluded target object.

\subsection{Perceptual Grounding: View Hallucination}
\label{sec:construction_angle}

\paragraph{Object selection.}
A single object is sampled from the same human-curated candidate pool used
for occlusion tasks. A confusable category map loaded from
pre-curated records (1000 categories and their confusable cateogires by GPT), for each category, which other
categories it is visually confusable with and under what viewing conditions,
providing the basis for the question posed to the evaluated model.

\paragraph{Placement.}
The object is placed on a randomly sampled support surface (table or floor)
via \texttt{OnTop} state sampling.

\paragraph{Agent initial viewpoint.}
The initial agent position is chosen from the set of azimuths at which the object is
most view-ambiguous.

\paragraph{Ground-truth label.}
The true category of the object, contrasted against its confusable
alternatives recorded in the metadata.

\subsection{Specular Reflection
: Reflection Authorization}
\label{sec:reflection_authorization}

\paragraph{Object selection.}
We first use GPT to obtain a list of placeable object categories from the BEHAVIOR-1K object inventory. We then filter out categories that are inherently reflective or structurally fixed, such as mirrors, walls, floors, doors, windows, and ceilings. From the remaining candidates, we retain only objects whose average-volume-derived equivalent spherical radius falls within \(0.02\)--\(0.20\) meter, ensuring that the selected objects are suitable for stable floor placement and mirror-reality reasoning. Finally, for each question, the target object category is randomly sampled using the scene name, room name, and question index as sampling context.

\paragraph{Mirror and initial-view setup.}
For each question, one mirror is first placed on a free floor position in the target room. The mirror is oriented to approximately face the room center, with a random yaw perturbation of up to \(30^\circ\). Its pose defines a geometric mirror plane, represented by a mirror center, a 2D normal direction, and a tangent axis. These parameters are used to compute whether an object should appear as a mirror reflection. After the mirror pose is fixed, the initial camera pose is selected based on the mirror placement: the camera is sampled from free floor positions on the front side of the mirror, preferring viewpoints approximately \(2.0\) m away from the mirror with a small lateral offset and oriented to look at the mirror center. A candidate initial view is accepted only if the line of sight from the camera to the mirror is not occluded by scene objects or non-traversable regions, ensuring that the mirror is clearly visible in the initial image.

\paragraph{Object placement.}
Each question places one real object on the floor. 
For positive examples, the object is sampled from positions whose reflected ray intersects the mirror plane within the mirror width and remains visible from the camera through the mirror. The object must lie at least \(1.0\) m from the mirror and must not be directly visible to the camera. 
For negative examples, the object is instead placed in a direct-only position: it is visible to the camera in the real world, but the geometric reflection test confirms that it is not visible through the mirror.

\paragraph{Ground-truth label.}
The ground-truth answer is a binary label indicating whether the queried object is seen through the mirror. Positive questions use the answer ``Yes'' when the object is mirror-visible but not directly visible. Negative questions use the answer ``No'' when the object is directly visible but has no valid mirror reflection.

\subsection{Specular Reflection
: Spatial Relationship}
\label{sec:spatial_relationship}

\paragraph{Object selection.}
We got the same object pool as \ref{sec:reflection_authorization}. For each question, two object categories are randomly sampled from this filtered pool.

\paragraph{Mirror and initial-view setup.}
We use the same mirror pose setup and initial-view setup as \ref{sec:reflection_authorization}.

\paragraph{Object placement.}
Each question places two real objects on the floor, denoted as object A and object B. Candidate object positions are sampled from free floor locations whose geometric mirror reflections are visible from the camera. For each candidate, the reflected ray from the camera through the mirror must intersect the mirror plane within the mirror width, remain inside the camera field of view, and pass both object-level and traversability-map occlusion checks. The object must also lie on the valid front side of the mirror and within the allowed mirror-depth range.

\paragraph{Ground-truth label.}
The ground-truth answer is a categorical label, either ``A'' or ``B''. It is computed by comparing the 3D Euclidean distance from the observation position to the real position of object A and object B. If object A is closer or tied, the answer is ``A''; otherwise, the answer is ``B''.

\subsection{Specular Reflection
: Scene Correspondence}

\paragraph{Object selection.}
We got the same object pool as \ref{sec:reflection_authorization}. For each question, two object categories are randomly sampled from this filtered pool.

\paragraph{Mirror and initial-view setup.}
We use the same mirror pose setup and initial-view setup as \ref{sec:reflection_authorization}.

\paragraph{Object placement.}
Each question places three identical real objects on the floor. The target object is sampled along the reflected viewing ray induced by the camera-to-mirror direction, at a distance of approximately \(1.0\)--\(2.0\) m from the mirror. The other two objects are placed on a line approximately perpendicular to this reflected ray, with offsets chosen so that the target object appears as the left, middle, or right object in the rendered image. All three placements must lie inside the valid floor region, avoid collisions with the mirror and scene blockers, and satisfy a minimum separation constraint. A small random yaw jitter is applied to each object to reduce visual degeneracy while preserving category and model identity. All candidate objects will be visible in our view.

\paragraph{Ground-truth label.}
The ground-truth answer is a categorical label from \(\{\text{left}, \text{middle}, \text{right}\}\). It is determined by the rendered-image horizontal position of the target real object, which is the object placed on the reflected mirror ray.

\subsection{Enumerative Perception
: Counting w Occlusion}
\label{sec:enumerative_counting_occlusion}

\paragraph{Object selection.}
We first use GPT to obtain a list of placeable common object pool from the BEHAVIOR-1K object inventory. Then we filter the pool by physical size: the representative model must have all bounding-box edges no larger than $0.25$ m, and cannot be so tiny that all three edges are below $0.05$ m.
This produces countable household objects that are small enough to be hidden by ordinary furniture, while remaining visible once the agent moves to a useful viewpoint.

\paragraph{Agent initial viewpoint.}
The agent position is taken from the input or from the loaded scene. 
If unavailable, we use the center of the selected floor as a default observer position. 
Images are rendered from the agent location with a camera height of about $1.25$ m above the floor and a downward pitch of $30^\circ$. 

\paragraph{Occluder selection and Object placement.}
We collect candidate occluders from the loaded OmniGibson room using their runtime bounding boxes and categories. 
We keep objects that are large enough to hide a small target object, such as cabinets, shelves, tables, sofas, counters, and similar furniture.

For each valid occluder, we place the count object behind it with respect to the agent's initial position. 
Specifically, we compute the direction from the agent to the occluder center, then place the target slightly beyond the occluder along this direction. 
We also require the line segment from the agent to the target to intersect the occluder's bounding box, so that the target is occluded from the initial viewpoint.

\paragraph{Ground-truth label.}
The ground-truth label is the number of target objects that are successfully placed in the scene. 
We give four number options for each question.

\subsection{Enumerative Perception
: Spatial Segmentation}
\label{sec:enumerative_segmentation}

\paragraph{Object selection.}
We remain the same object pool in \ref{sec:enumerative_counting_occlusion}.

\paragraph{Divider selection and Object placement.}
For this task, we use chair-like objects as natural visual dividers, including chairs, armchairs, stools, and swivel chairs. 
Structural or non-blocking objects such as floors, walls, doors, windows, lights, and pictures are excluded.

For each valid divider, we place the count object at the horizontal center of the divider's bounding box, on the floor underneath or within the divider region. 
This creates a case where the object is spatially separated or visually broken up by the surrounding furniture structure. 

\paragraph{Agent initial viewpoint.}
The same agent initial view as \ref{sec:enumerative_counting_occlusion}.

\paragraph{Ground-truth label.}
The same ground-truth label as \ref{sec:enumerative_counting_occlusion}.

\subsection{Enumerative Perception
: Merged Observation}
\label{sec:enumerative_merge}

\paragraph{Object selection.}
We remain the same object pool in \ref{sec:enumerative_counting_occlusion}.

\paragraph{Object placement.}
We create a visually merged counting case by placing multiple instances of the same count object in a compact cluster. 
We first sample a free floor position from the room, ensuring that it lies inside the valid room region and does not collide with blocking objects.

Starting from this base position, we add the remaining target objects nearby with small random offsets. 
The offsets are constrained so that neighboring objects stay very close to each other, and in some cases objects are stacked on top of nearby ones. 
This produces a group where individual instances may visually merge into one another from the agent's viewpoint.

\paragraph{Agent initial viewpoint.}
The same agent initial view as \ref{sec:enumerative_counting_occlusion}.

\paragraph{Ground-truth label.}
The same ground-truth label as \ref{sec:enumerative_counting_occlusion}.

\subsection{Enumerative Perception
: Category Ambiguity}
\label{enumerative_ambiguity}

\paragraph{Object selection.}
We first maintain the same objest pool as \ref{sec:enumerative_counting_occlusion}.
To create semantic ambiguity, we compute CLIP similarity scores for all object pairs in the pool. For each object category, we keep its top-3 most similar categories as candidate confusers. For each question, we first sample a counting object category. Then, for each counting object added to the scene, we also add one confuser object, randomly selected from its top-3 candidate confusers.

\paragraph{Object placement.}
We first sample valid free-floor positions inside the room, requiring each position to lie on the selected floor, avoid collisions with blocking objects, and remain sufficiently far from the agent. 
For each candidate, we place one target object at the sampled free position. 
We then search for a nearby valid position for the confuser object using small offsets around the target location. 
The confuser is placed close to the target object, but still at a collision-free floor position. 

\paragraph{Agent initial viewpoint.}
The same agent initial view as \ref{sec:enumerative_counting_occlusion}.

\paragraph{Ground-truth label.}
The same ground-truth label as \ref{sec:enumerative_counting_occlusion}.

\subsection{Enumerative Perception
: Structural Enclosure}
\label{sec:enumerative_enclosure}
\paragraph{Object selection.}
We first maintain the same objest pool as \ref{sec:enumerative_counting_occlusion}. For container object, we fixed them with several types of box, to make sure they all can open and close.

\paragraph{Object placement.}
Each container is instantiated in the room and placed either on a valid support surface, such as a table, or on a free floor position if no suitable support is available.

We then randomly choose a subset of these containers to contain the target objects. 
For each selected container, the target object position is set near the center of the container's bounding box, and the container is kept closed when possible. 
Containers not selected by the random sampling remain empty. 

\paragraph{Agent initial viewpoint.}
The same agent initial view as \ref{sec:enumerative_counting_occlusion}.

\paragraph{Ground-truth label.}
The same ground-truth label as \ref{sec:enumerative_counting_occlusion}.

\subsection{Enumerative Perception
: Illumination Variability}

\paragraph{Object selection.}
We first maintain the same objest pool as \ref{sec:enumerative_counting_occlusion}. 

\paragraph{Object placement.}

For the light-change task, we first sample valid free-floor positions inside the room. 
Target objects are placed at these collision-free positions. 
We then adjust the ambient light intensity and add directional lights to selected objects, creating views with different illumination conditions while keeping the object positions fixed.

\paragraph{Agent initial viewpoint.}
The same agent initial view as \ref{sec:enumerative_counting_occlusion}.

\paragraph{Ground-truth label.}
The same ground-truth label as \ref{sec:enumerative_counting_occlusion}.

\subsection{Cognitive Mapping
: Connectivity}
\label{sec:cognitive_connectivity}

\paragraph{Agent initial viewpoint.}
The connect task selects one navigable render point for each room from the traversability map, preferring free cells with high clearance and near the room center. From this point, it captures eight evenly spaced room reference views. Each question then attaches the views of its required rooms, and uses the first available room view as the main visual context.

\paragraph{Question Generation.}
The script extracts room regions from the scene segmentation map and samples navigable candidate points for each room. It then generates question: whether two rooms are connected. For each candidate room pair, it calls \texttt{scene.get\_shortest\_path} between sampled points and projects the returned world path back to a room sequence using the segmentation map.

\paragraph{Ground-truth label.}
For pair connectivity, the label is \texttt{Yes} if a valid shortest path exists between the two rooms, and \texttt{No} otherwise.

\subsection{Cognitive Mapping
: Traversable Passage}
\label{sec:cognitive_traversable}

\paragraph{Agent initial viewpoint.}
The same as \ref{sec:cognitive_connectivity}.

\paragraph{Question Generation.}
The script extracts room regions from the scene segmentation map and samples navigable candidate points for each room. It then generates question: whether the shortest path between two rooms passes through a third room.  For each candidate room pair, it calls \texttt{scene.get\_shortest\_path} between sampled points and projects the returned world path back to a room sequence using the segmentation map.

\paragraph{Ground-truth label.}
For shortest-path-via-region questions, the label is \texttt{Yes} if the queried via-room appears inside the projected shortest room sequence, excluding the source and target rooms; otherwise it is \texttt{No}.

\subsection{Cognitive Mapping
: Regional Boundary}
\label{sec:cognitive_regional}

\paragraph{Object Selection.}
The script collects objects in the scene after filtering out robots, structural/background categories, doors, floors, walls, ceilings, mirrors, and oversized objects. Only objects referenced by generated questions are rendered, with four object-centric views captured around the object.

\paragraph{Agent initial viewpoint.}
The same as \ref{sec:cognitive_connectivity}.

\paragraph{Question Generation.}
The script extracts room regions from the segmentation map. It then assigns valid scene objects to regions and generates three question types: whether an object belongs to a region, whether two objects are in the same region, and which of two regions an object is closer to.

\paragraph{Ground-truth label.}
For object-in-region questions, the label is \texttt{Yes} if the queried region matches the object's assigned region, and \texttt{No} otherwise. For same-region questions, the label is \texttt{Yes} if the two objects have the same assigned region. For closer-region questions, the answer is the region with the smaller distance from the object center to the region bbox, with region-center distance used as a tiebreaker.

\subsection{Cognitive Mapping
: Long-Term Navigation}
\label{sec:cognitive_longterm}

\paragraph{Agent initial viewpoint.}
The same as \ref{sec:cognitive_connectivity}.

\paragraph{Question Generation.}
The script extracts room regions from the segmentation map. It then searches reachable room pairs with \texttt{scene.get\_shortest\_path}, ranks them by path distance, and keeps the farthest connected pairs. For each selected pair, it generates two question types: \texttt{navigation\_actions}, which asks for turn/action choices along the route, and \texttt{navigation\_regions}, which asks for the next region to explore at each step.

\paragraph{Ground-truth label.}
The ground truth is derived from the selected shortest path and its projected region sequence. For \texttt{navigation\_regions}, each answer is the next region in the path sequence. For \texttt{navigation\_actions}, go-to-region steps are fixed by the same sequence, while turn-choice answers are computed from the change in heading between consecutive regions and labeled as \texttt{turn left}, \texttt{turn right}, or \texttt{turn back}.





\appendix



\end{document}